\documentclass{article}

\usepackage{microtype}
\usepackage{graphicx}
\usepackage{subcaption}
\usepackage{booktabs} % for professional tables

\usepackage{hyperref}

\usepackage[preprint]{icml2026}

\usepackage{amsmath}
\usepackage{amssymb}
\usepackage{mathtools}
\usepackage{amsthm}

\usepackage[capitalize,noabbrev]{cleveref}

\theoremstyle{plain}

\theoremstyle{definition}

\theoremstyle{remark}

\usepackage[textsize=tiny]{todonotes}

\icmltitlerunning{STELLAR: Scaling 3D Perception Large Models for Autonomous Driving}

\usepackage{booktabs} % For professional looking tables
\usepackage{multirow} % For multi-row cells
\usepackage{graphicx} % For resizing columns
\usepackage[table]{xcolor} % For row colors
\usepackage[normalem]{ulem}
\newcommand{\ourmethod}{STELLAR}

\begin{document}

\twocolumn[
  \icmltitle{STELLAR: Scaling 3D Perception Large Models for Autonomous Driving}



  \icmlsetsymbol{equal}{*}

  \begin{icmlauthorlist}
    \icmlauthor{Yingwei Li}{equal,waymo}
    \icmlauthor{Xin Huang}{equal,waymo}
    \icmlauthor{Yang Liu}{equal,waymo}
    \icmlauthor{Yang Fu}{waymo,ucsd}
    \icmlauthor{Alex Zihao Zhu}{waymo}
    \icmlauthor{Chen Song}{waymo}
    \icmlauthor{Junwen Yao}{waymo}
    \icmlauthor{Anant Subramanian}{waymo}
    \icmlauthor{Hao Xiang}{waymo}
    \icmlauthor{Weijing Shi}{waymo}
    \icmlauthor{Yuliang Zou}{waymo}
    \icmlauthor{Tom Hoddes}{waymo}
    \icmlauthor{Zhaoqi Leng}{waymo}
    \icmlauthor{Govind Thattai}{waymo}
    \icmlauthor{Dragomir Anguelov}{waymo}
    \icmlauthor{Mingxing Tan}{waymo}
  \end{icmlauthorlist}

  \icmlaffiliation{waymo}{Waymo, Mountain View, CA, USA}
  \icmlaffiliation{ucsd}{UCSD, La Jolla, CA, USA}

  \icmlcorrespondingauthor{Yingwei Li}{ywli@waymo.com}
  \icmlcorrespondingauthor{Xin Huang}{xchuang@waymo.com}
  \icmlcorrespondingauthor{Mingxing Tan}{tanmingxing@waymo.com}

  \icmlkeywords{Autonomous Driving, Perception, ICML}

  \vskip 0.2in
]



\printAffiliationsAndNotice{\icmlEqualContribution}

\begin{abstract}
Model scaling has demonstrated remarkable success through large-scale training on diverse datasets.
It remains an open question whether the same paradigm would apply to autonomous driving perception systems due to unique challenges, such as fusing heterogeneous sensor data and the need for sophisticated 3D spatial understanding. 
To bridge this gap, we present a comprehensive study on systematically analyzing the impact of scale on these systems.
We develop our STELLAR model based on Sparse Window Transformer, by extending the input modalities to include LiDAR, radar, camera, and map prior.
We train the model on a large-scale dataset of 50 million driving examples with up to 500 million parameters.
Our large-scale experiments reveal empirical scaling trends that connect model performance to model size, data, and compute.
The resulting model establishes a new state-of-the-art on the Waymo Open Dataset challenge, outperforming prior arts by a large margin.
Our work demonstrates that large-scale training is a highly promising path for advancing the capabilities of perception models for autonomous driving.
% \vspace{-2mm}
\end{abstract}    
\section{Introduction}
\label{sec:intro}

% Describe perception for autonomous driving.
Perception is a fundamental task for autonomous driving, responsible for interpreting the surrounding environment using data from sensors such as cameras, LiDAR, and radar.
Recent breakthroughs have significantly enhanced these perception models, leading to more accurate 3D bounding box detection~\cite{zhang2023hednet,zhang2024safdnet,huang2025vadet,agro2025mad} and a comprehensive understanding of the scene through segmentation~\cite{wu2024point,sun2024uni}.
These enhanced capabilities are foundational to enabling safer and more robust driving in complex environments~\cite{schreier2023offline,hu2023planning}.

\begin{figure}[t]
    \centering
    \includegraphics[width=1\linewidth]{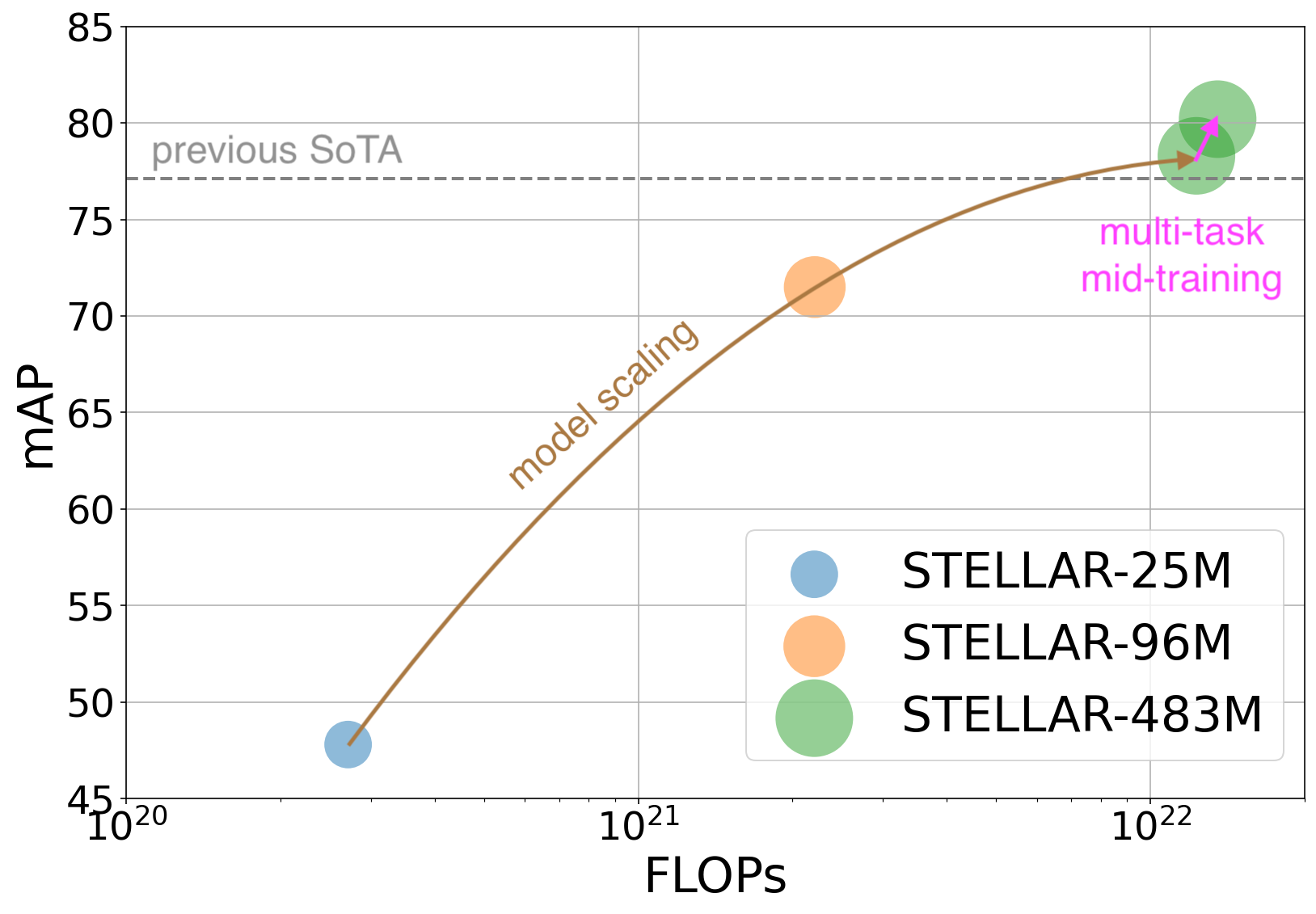}
    % \vspace{-0.05in}
    \caption{\ourmethod{} achieves better 3D detection performance through scaling model parameters and multi-task mid-training on high quality driving data, measured by average L2 APH on the Waymo Open Dataset validation set. The dashed horizontal line represents previous state-of-the-art using up to 4 temporal frames.}
    \label{fig:teaser}
    % \vspace{-2mm}
\end{figure}

% Describe scaling in general.
Recently, large foundation models have represented a significant paradigm shift in artificial intelligence, driven by the principle of training at an unprecedented scale. 
By dramatically increasing model parameters and leveraging vast, internet-scale datasets, these models have achieved state-of-the-art performance and unlocked emergent capabilities across both language~\cite{brown2020language,kaplan2020scaling,hoffmann2022training} and vision~\cite{achiam2023gpt,team2023gemini} domains.

% Describe scaling for perception models.
It remains an open question whether these scaling paradigms apply to 3D perception for autonomous driving, due to several unique and formidable challenges.
% Inspired by this paradigm shift, we aim to apply similar scaling principles to perception models for autonomous driving, which presents several unique challenges.
First, unlike existing foundation models that process language text or image inputs, driving models typically ingest and fuse heterogeneous data from multiple sensors, combining multi-modal features from LiDAR point clouds, multi-view camera images, radar images, and map prior. Developing architectures to effectively account for multi-modal sensor fusion is crucial to model scaling.
Second, processing these dense, high-resolution sensor inputs over long ranges imposes significant computational challenges as model size increases.
Third, driving perception demands a sophisticated 3D spatial understanding of the world. This requirement for precise geometric prediction within dynamic environments distinguishes it from general-purpose foundation models, which predominantly operate on 2D images or textual data.
Finally, the training paradigms differ significantly. While language models leverage self-supervised learning on unlabeled data, perception models typically rely on high-quality, manually-annotated labels. This costly process creates a major bottleneck for data scaling.

% Describe our contributions.
In this paper, we present a comprehensive study on scaling a multi-modal perception Transformer model for autonomous driving across both model and data, as shown in Fig.~\ref{fig:teaser}.
Our main contributions are threefold:
\begin{itemize}
    \item We develop a scalable \ourmethod{} model that jointly process multimodal data, from LiDAR, radar, and camera sensors, using a unified sparse window Transformer backbone~\cite{sun2022swformer}.
    \item We scale up our  models up to 500 million parameters and train them with 50 million examples from our propriety driving logs. Both the model size and data size are orders of magnitude larger than prior art.
    We identify and characterize the underlying scaling properties that uncover performance gains related to model size, data volume, and compute.
    \item Our best \ourmethod{} model establishes a new state-of-the-art on the competitive Waymo Open Dataset, outperforming prior arts by a large margin.
\end{itemize}
\section{Related Work}
\label{sec:related_work}

\subsection{Perception for Autonomous Driving}
Perception is a fundamental capability for autonomous systems, responsible for processing raw sensor inputs to interpret the driving environment.
Substantial progress has advanced performance across diverse perception tasks, including 3D object detection~\cite{zhang2023hednet,zhang2024safdnet,huang2025vadet,agro2025mad}, segmentation~\cite{wu2024point,sun2024uni}, and occupancy prediction~\cite{wei2023surroundocc,tian2023occ3d}.

Despite these successes, it remains unclear how effectively autonomous driving perception models scale with increased data and model size. This gap persists largely due to the significant challenges of scaling architectures that process dense multi-modal inputs, alongside the prohibitive cost of collecting and labeling driving data at a massive scale. To address this uncertainty, our work provides a comprehensive empirical study of scaling for perception, specifically 3D detection, in autonomous driving. We train models ranging up to 500 million parameters on a dataset of over 50 million driving examples.

\subsection{3D Object Detection}
3D object detection serves as one of the most critical perception tasks in autonomous driving, with modern approaches including LiDAR and camera-based paradigms. 
LiDAR detectors~\cite{zhou2018voxelnet,sun2022swformer,zhang2024safdnet,huang2025vadet} have achieved dominant performance by leveraging sparse voxelization and efficient backbone architectures. Concurrently, camera-based detection has undergone a pivotal shift from perspective-view to bird's-eye-view (BEV) representations~\cite{philion2020lift,huang2021bevdet,li2024bevformer}.
To enable better scalability, our approach synthesizes these paradigms: we extend the sparse, voxel-based SWFormer backbone~\cite{sun2022swformer} to integrate camera features, by projecting camera features into a unified BEV space following~\cite{philion2020lift}.

\subsection{Foundation Models for Autonomous Driving}
Foundation models trained on web-scale data have established a significant paradigm shift in artificial intelligence, achieving state-of-the-art results across language~\cite{radford2018improving,devlin2019bert} and vision domains~\cite{radford2021learning,caron2021emerging}. These breakthroughs are underpinned by established scaling properties~\cite{kaplan2020scaling,hoffmann2022training} that quantify the relationship between model size, dataset volume, and performance.

Within autonomous driving, self-supervised learning has emerged as a dominant strategy for leveraging vast amount of unlabeled LiDAR data. Early work~\cite{hess2023masked} demonstrates that geometric reconstruction tasks can yield robust features without manual labels. 
Subsequent research has introduced more complex representations, utilizing unsupervised occupancy fields~\cite{agro2024uno} and 3D volumetric rendering~\cite{yang2024unipad} as supervision signals. More recent paradigms~\cite{ljungbergh2025gasp,wozniak2025s3pt} further unify geometric and semantic signals to improve representations for downstream perception tasks.

While these paradigms primarily focus on the design of proxy tasks to learn from unlabeled data,~\ourmethod{} investigates a different axis: the impact of scaling auto-labeled and supervised data on 3D perception. While prior work has successfully verified scaling laws for behavior prediction and planning models~\cite{huang2025drivegpt,baniodeh2025scaling}, these systems typically operate on sparse inputs (e.g., map coordinates and trajectories). 
Scaling perception models, however, presents a distinct and greater computational challenge. This difficulty stems from processing dense, high-dimensional sensor inputs like camera data and massive LiDAR point clouds, leaving perception scalability an open question due to the challenges of scaling raw data inputs and their corresponding training targets. 

In this work, we address this gap by presenting the first empirical scaling study for autonomous driving perception. We establish this study by training a transformer-based, multi-modal model while scaling both its parameters and the training dataset to an unprecedented scale.
\begin{table}[t!]
% \vspace{0em}
\centering
\begin{tabular}{l@{\hspace{2pt}}c@{\hspace{2pt}}c}
\toprule
\textbf{Statistics} & \textbf{Ours} & \textbf{Prior Work} \\
\midrule
Run segments     & 59M          & 1M{\footnotesize\cite{yaak2025l2d}}         \\
Hours of driving       & 169K          & 5K{\footnotesize\cite{yaak2025l2d}}          \\
Camera images     & 47.2B & 2.7M{\footnotesize\cite{wilson2023argoverse}} \\
LiDAR frames  & 6B          & 6M{\footnotesize\cite{wilson2023argoverse}}          \\
3D bounding boxes  & 431B          & 5B{\footnotesize\cite{karnchanachari2024towards}}          \\
\bottomrule
\end{tabular}
\caption{We train~\ourmethod{} on a large-scale 3D perception driving dataset, which includes multi-modal input features at a scale exceeding existing literature.}
\label{tab:data_stats}
\vspace{-2em}
\end{table}

\section{Data}
\label{sec:data}

We introduce a large-scale driving dataset designed to study the scaling properties of perception models in autonomous driving. Collected from millions of miles of high-quality human and autonomous driving demonstrations, the dataset captures extensive spatiotemporal diversity. It spans more than 10 U.S. cities, covering urban, suburban, freeway, and indoor (e.g., parking garage) environments across various times of day and adverse weather conditions.

The dataset comprises over 50M training examples. Each example is derived from a unique 10-second driving segment discretized at 10Hz, consistent with standard benchmarks such as Waymo Open~\cite{sun2020scalability} and nuScenes~\cite{caesar2020nuscenes}. The sensor suite features multi-modal data including LiDAR, camera, and radar. Additionally, we follow~\cite{pfister2000surfels,yang2020surfelgan} to construct a surfel map prior by estimating mean coordinates, surface normals, and color from aligned LiDAR points and camera pixels. To construct a single training example from the 100 frames available in each segment, we randomly sample 4 frames (1 current and 3 historical). Key statistics and comparisons to prior work are detailed in Tab.~\ref{tab:data_stats}.

For supervision, our dataset provides 7-DoF (center$_\text{x}$, center$_\text{y}$, center$_\text{z}$, length, width, height, heading) 3D bounding box labels for multiple object categories, comprising both manual annotations and auto-generated pseudo-labels. Due to the complexity of manual annotation at scale, we employ an off-board autolabeler based on~\cite{li2023modar}. This model, which leverages long-term temporal information (past and future frames) and takes multi-modality sensor data as inputs, was trained on our high-quality, manually-annotated subset. The trained autolabeler was then used to generate pseudo-labels for the remainder of our dataset.

\section{\ourmethod}

We present \ourmethod{}, a simple and scalable perception model designed to operate across multiple dimensions: input modalities, temporal context, perception tasks, model size, and perception ranges. The high-level model architecture is shown in Fig.~\ref{fig:motivating_example}. The remainder of this section will elaborate on how \ourmethod{} achieves scalability across dimensions.

\subsection{Multi-Modal and Multi-Frame Sensor Inputs}
\begin{figure}
    \centering
    \includegraphics[width=1.0\linewidth]{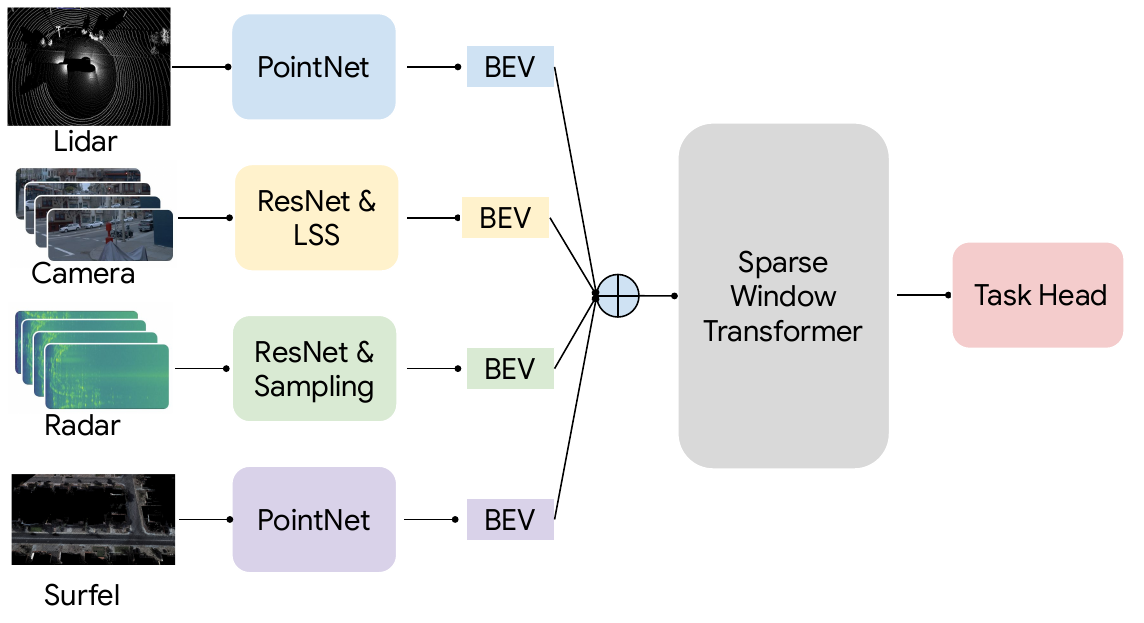}
    \caption{Overview of~\ourmethod{}, a multi-modal perception model. The model projects LiDAR, radar, and surfel inputs directly into a bird's-eye-view (BEV) representation, while camera features are mapped to BEV via a lift-splat-shoot (LSS) transformation. These features are subsequently concatenated and processed by a sparse window transformer backbone. Task-specific heads are applied to the unified BEV features to produce the final outputs.}
    \label{fig:motivating_example}
    \vspace{-2em}
\end{figure}

Our model takes four modalities: LiDAR point cloud, camera images, radar polar images, and surfel map prior, as described in Sec.~\ref{sec:data}. We use modality specific encoders to construct bird's-eye-view feature maps. The LiDAR encoder, following~\cite{sun2022swformer}, uses dynamic voxelization~\cite{zhou2020end} and a PointNet-style embedding model~\cite{qi2017pointnet,lang2019pointpillars} to generate sparse voxel features. We introduce an extra projection layer to align LiDAR features with camera and radar modalities. For camera images, a ResNet~\cite{he2016deep} and a lift-splat-shoot (LSS) module~\cite{philion2020lift} are used to produce dense image features in bird's-eye-view (BEV). A separate ResNet is employed to encode radar polar images, and their resulting feature maps are resampled into Cartesian coordinates. The surfel map prior shares the same format as LiDAR points, and is then fed into a surfel encoder with the same architecture of the LiDAR encoder~\cite{fu2026scene}.  These encoded features are then fused into a unified BEV feature map through concatenation, similar to BevFusion~\cite{bevfusion23icra}, as shown in Fig.~\ref{fig:motivating_example}. 

To further enhance capability, we aggregate LiDAR features over a temporal window following~\cite{luo2018fast}, which allows the model to reconstruct geometries of partially occluded objects from varying viewpoints and utilize temporal cues to refine object heading estimation.

\subsection{Perception Tasks}
\label{sec:task_scaling}
The BEV features are fed into a multi-scale SWFormer~\cite{sun2022swformer} backbone that consists of a multi-scale feature extractor and a multi-scale feature fuser. The backbone's output is then passed to three distinct heads for 3D detection, occupancy prediction, and roadgraph prediction tasks.

The 3D detection head extends SWFormer's approach, performing detection in two stages. First, it refines foreground and background voxels via segmentation and voxel diffusion. Second, a CenterNet-style~\cite{duan2019centernet} detection head produces the final classification and box regression outputs.

The occupancy head and roadgraph head each employ a series of convolutional layers to refine backbone features and subsequently produce per-voxel logits. For occupancy prediction, these logits are passed to an MLP to generate a per-voxel semantic classification. In contrast, for roadgraph prediction, the logits are used to predict predefined geometric targets, such as lanes and road boundaries.

\subsection{Scalable Model Size} 
Although the model processes different modalities and performs various tasks, the computational load is primarily concentrated in the transformer backbone. For instance, in our largest model, the backbone accounts for over 90\% of the total parameters. 
This centralized architectural design simplifies the scaling process and requires less hyperparameter tuning. 
To implement this, we scale the model size by increasing both its width (hidden dimension size, feed-forward ratio) and depth (number of layers), following existing scaling literature~\cite{kaplan2020scaling,hoffmann2022training,petty2024impact}.

\subsection{Scalable Perception Range} 
In practical applications, perception models must support different perception ranges depending on the operational scenario. For example, freeway driving typically requires a larger perception range than driving on surface streets. By virtue of its transformer-style architecture and sparse representation, our model supports training at one range and inference at a different range without requiring any changes to the model weights. While this paper follows the SWFormer~\cite{sun2022swformer} range settings for the Waymo Open Dataset, our design is inherently scalable to larger or smaller ranges as dictated by specific  requirements.

\subsection{Scaling Down with Distillation} 
We also study the process of distillation, which can improve the accuracy of smaller and cheaper model variants using the scaled up \ourmethod{} models as teachers. This approach leverages teacher predictions directly from raw CenterNet heatmaps as training targets for the student heatmaps. We utilize the standard CenterNet loss framework, specifically adapting the focal loss to accommodate continuous targets.

\section{Training Recipe}
\label{sec:recipe}

We employ a three-stage training strategy for \ourmethod{}: the model is initially pre-trained on the $50$ million pre-training dataset (Sec.~\ref{sec:data}), subsequently trained on $600K$ human-annotated segments, and finally fine-tuned on $798$ WOD segments.

During pre-training, we employ the LAMB optimizer~\cite{you2019large} with a constant learning rate of $1e^{-2}$ and a global batch size of $1024$.
Following~\cite{he2019rethinking,ibrahim2024simple,hagele2024scaling}, we utilize a constant learning rate schedule to enable evaluation and mid-training at various checkpoints without the need for retraining.
Training proceeds for up to $50K$ steps, approximately equivalent to one epoch over the dataset.
To minimize training on temporally correlated frames and maximize data diversity, we sample our batches by selecting one random frame plus three history frames per driving segment.
We supervise solely on positive, high-confidence auto-labels (i.e., confidence score $\geq 0.3$). We set the DropPath ratio~\cite{huang2016deep} to $0.5$ and weight decay to $1e^{-3}$. For simplicity, we focus exclusively on the 3D detection task and utilize standard LiDAR, camera, and radar inputs, due to the complexity of generating surfel inputs at this scale.

During mid-training, we retain the same optimizer but adopt a cosine decay learning rate schedule, starting with an initial learning rate of $1e^{-4}$. We train for $20K$ steps with a batch size of $256$. In this phase, we introduce additional internal dense tasks, such as occupancy prediction and roadgraph prediction, and incorporate surfels as an extra input modality generated from LiDAR and camera inputs, as described in Sec.~\ref{sec:data}.

Finally, we finetuned on the WOD training set for the 3D detection task, using a cosine decay learning rate schedule with an initial learning rate of $3e^{-5}$. We train for $10K$ steps with a global batch size of $256$. We applied data augmentation as described in~\cite{sun2022swformer}.
\section{Scaling}
This section outlines scaling methodology and reports empirical results during pre-training. We scale towards 500M parameters and 50M driving examples, representing a cumulative computational budget exceeding 10 ZettaFLOPs.

Notably, to address high device memory pressure from model scaling, we employed gradient checkpointing~\cite{chen2016training,bulatov2018fitting}, which discards intermediate activations within PointNet and SWFormer during the forward pass and recomputes them for backpropagation.

\begin{table}[t!]
\small
\centering
\begin{tabular}{c|ccc}
\toprule
Model Size & Width & Feed-forward Ratio & Layers \\
\midrule
25M & 128 & 2 & [2, 3, 2, 3, 2]          \\
96M & 256 & 8 & [4, 6, 4, 6, 4]          \\
251M & 384 & 10 & [5, 7, 5, 7, 5]          \\
364M & 384 & 15 & [5, 8, 5, 8, 5]          \\
483M & 384 & 18 & [6, 9, 6, 9, 6]          \\
\bottomrule
\end{tabular}
\caption{Overview of~\ourmethod{} configurations at different sizes, by modifying key parameters in the Sparse Window Transformer.}
\label{tab:model_scaling}
% \vspace{-2em}
\end{table}

\begin{figure}[t]
    \centering
    \includegraphics[width=1\linewidth]{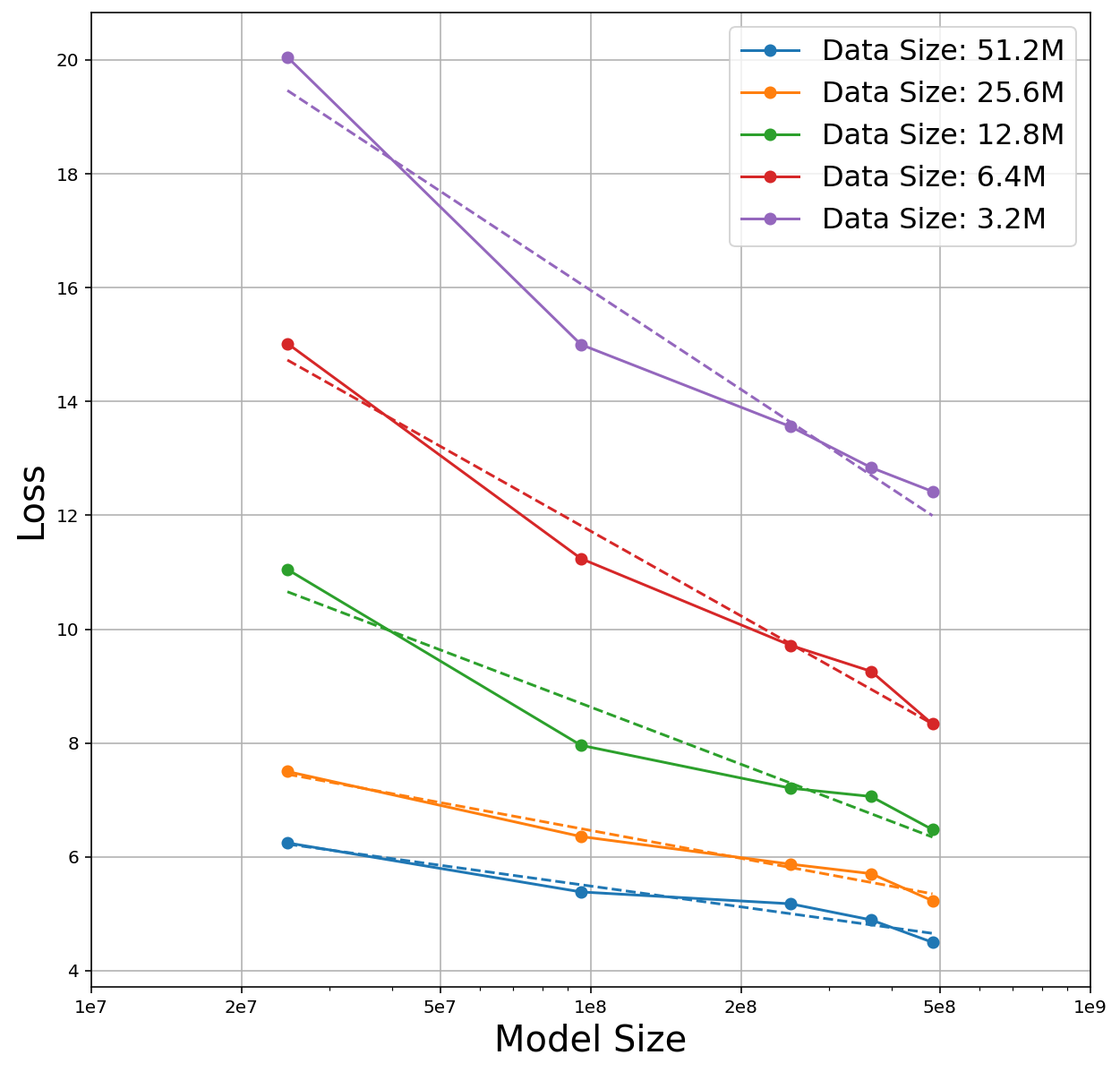}
    \caption{Model scaling curves. Final loss consistently decreases as model parameter size increases. Log-linear fits are overlaid for each dataset size to illustrate the scaling trend.}
    \label{fig:model_scaling}
    % \vspace{-2em}
\end{figure}

\begin{figure}[t]
    \centering
    \includegraphics[width=1\linewidth]{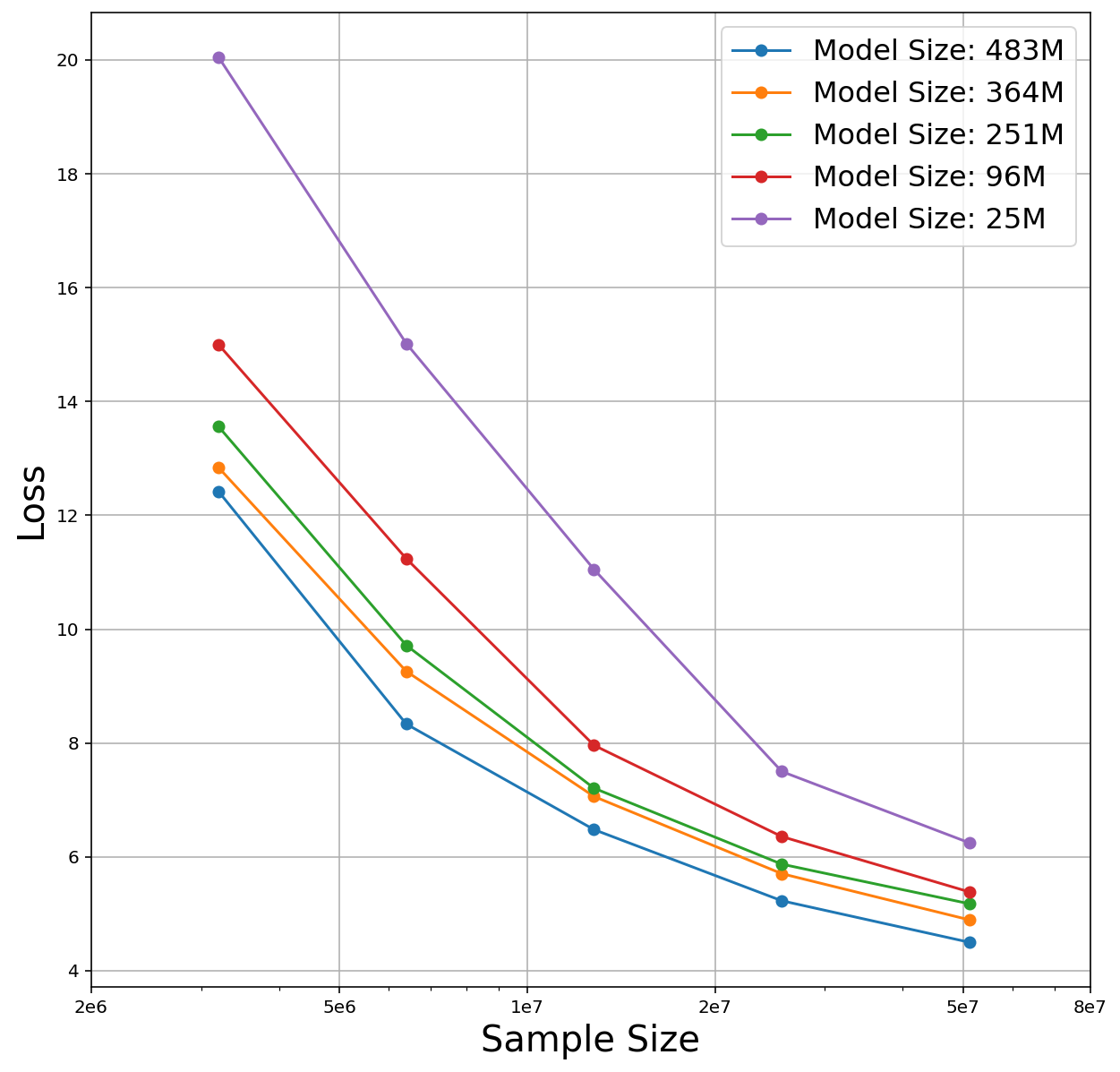}
    \caption{Data scaling curves. Increasing the training example size consistently reduces the final loss for all model sizes.}
    \label{fig:data_scaling}
    % \vspace{-2em}
\end{figure}

\subsection{Model Scaling}
We first scale~\ourmethod{} at different model sizes, by varying the transformer parameters, including hidden dimension size, feed-forward ratio, and number of layers, as shown in Tab.~\ref{tab:model_scaling}.

Figure~\ref{fig:model_scaling} illustrates model performance as a function of parameter count across varying dataset sizes.
We observe two trends. First, for any fixed dataset size, increasing the model's parameter count consistently reduces the final loss. Second, the magnitude of this gain from model scaling is highly dependent on the dataset size. The loss curves are steepest for smaller datasets (e.g., 3.2M examples) and become progressively flatter as the dataset size increases (e.g., 51.2M examples).

We add a log-linear fit for each curve in dashed lines. Unlike the strong, log-linear scaling laws commonly observed in LLM literature, our observed scaling trend exhibits diminishing returns. We attribute this to two factors. First, our model is a complex, multi-component architecture (e.g., with various encoders), not a single, homogeneous transformer decoder. Second, we primarily scale the transformer backbone while other components remain fixed. This fixed-size parameter overhead distorts the scaling curve and masks the true scaling behavior of the transformer component.

\subsection{Data Scaling}

Fig.~\ref{fig:data_scaling} illustrates the impact of data scaling on models of varying sizes. We observe a consistent trend that increasing the training example size monotonically reduces the loss for all models, though this benefit exhibits diminishing returns as the loss curves flatten.

Similar to our model scaling findings, we do not observe the strong log-linear scaling laws often reported in LLM literature. We attribute this to data quality being bounded by the qualities of auto-labels and the intrinsic redundancy of driving data, even across distinct segments. We leave further investigation of these factors to future research.

\begin{table*}[t]
  \centering
  \resizebox{\textwidth}{!}{%
  \begin{tabular}{@{}l cc cc cc cc cc cc cc cc@{}}
    \toprule
    \multirow{2}{*}{Method} & \multicolumn{2}{c}{Overall L1} & \multicolumn{2}{c}{Overall L2} & \multicolumn{2}{c}{Vehicle L1} & \multicolumn{2}{c}{Vehicle L2} & \multicolumn{2}{c}{Pedestrian L1} & \multicolumn{2}{c}{Pedestrian L2} & \multicolumn{2}{c}{Cyclist L1} & \multicolumn{2}{c}{Cyclist L2} \\
    & AP & APH & AP & \underline{APH} & AP & APH & AP & APH & AP & APH & AP & APH & AP & APH & AP & APH \\
    \midrule
    PointPillars~\cite{lang2019pointpillars} & 69.0 & 63.5 & 62.8 & 57.8 & 72.1 & 71.5 & 63.6 & 63.1 & 70.6 & 56.7 & 62.8 & 50.3 & 64.4 & 62.3 & 61.9 & 59.9 \\
    SST~\cite{fan2022embracing} & 74.5 & 71.0 & 67.8 & 64.6 & 74.2 & 73.8 & 65.5 & 65.1 & 78.7 & 69.6 & 70.0 & 61.7 & 70.7 & 69.6 & 68.0 & 66.9 \\
    CenterFormer~\cite{zhou2022centerformer} & 75.6 & 73.2 & 71.4 & 69.1 & 75.0 & 74.4 & 69.9 & 69.4 & 78.0 & 72.4 & 73.1 & 67.7 & 73.8 & 72.7 & 71.3 & 70.2\\
    SWFormer~\cite{sun2022swformer} & - & - & - & - & 77.8 & 77.3 & 69.2 & 68.8 & 80.9 & 72.7 & 72.5 & 64.9 & - & - & - & - \\
    % MoDAR~\cite{li2023modar} & - & - & - & - & 84.5 & 84.0 & 77.5 & 77.0 & 86.3 & 82.5 & 79.5 & 75.8 & - & - & - & - \\
    CenterPoint~\cite{yin2021center} & 77.5 & 75.8 & 71.7 & 70.1 & 76.4 & 75.9 & 68.7 & 68.2 & 79.2 & 75.6 & 71.9 & 68.5 & 76.8 & 75.9 & 74.4 & 73.5 \\
    FSDv1~\cite{fan2022fully} & 79.6 & 77.4 & 72.9 & 70.8 & 79.2 & 78.8 & 70.5 & 70.1 & 82.6 & 77.3 & 73.9 & 69.1 & 77.1 & 76.0 & 74.4 & 73.3 \\
    FSDv2~\cite{fan2024fsd} & 81.8 & 79.5 & 75.6 & 73.5 & 79.8 & 79.3 & 71.4 & 71.0  & 84.8 & 79.7 & 77.4 & 72.5 & 80.7 & 79.6 & 77.9 & 76.8 \\
    HEDNet 4f~\cite{zhang2023hednet} & 83.6 & 82.3 & 78.1 & 76.8 & 82.4 & 81.9 & 75.1 & 74.6 & 86.3 & 83.6 & 79.4 & 76.8 & 82.2 & 81.4 & 79.9 & 79.1 \\
    SAFDNet 4f~\cite{zhang2024safdnet} & 83.9 & 82.6 & 78.4 & 77.1 & 82.8 & 82.3 & 75.4 & 74.9 & 86.8 & 84.2 & 80.1 & 77.5 & 82.0 & 81.1 & 79.6 & 78.8 \\
    % MAD~\cite{agro2025mad} & 85.8 & 84.2 & 81.0 & 79.4 & 84.2 & 83.6 & 77.4 & 76.8 & 87.9 & 85.4 & 82.2 & 79.7 & 85.2 & 83.7 & 83.3 & 81.7 \\
    \midrule
    % \ourmethod{}-500M (10M data) & - & - & \textbf{82.5} & \textbf{80.9} & - & - & \textbf{85.4} & \textbf{84.9} & - & - & 79.9 & 76.4 & - & - & 82.3 & 81.4 \\
    % (100M data @ 100K ckpt)
    % \ourmethod{}-500M & - & - & 82.3 & 80.6 & - & - & 84.3 & 83.7 & - & - & 80.2 & 76.7 & - & - & 82.6 & 81.5 \\
    % (100M data @ 400K ckpt)
    % \ourmethod{}-500M & - & - & \textbf{83.4} & \textbf{81.9} & - & - & \textbf{85.9} & \textbf{85.3} & - & - & \textbf{81.0} & \textbf{77.7} & - & - & \textbf{83.5} & \textbf{82.6} \\
    % https://waymo.com/open/challenges/detection-3d/results/6f5c8ccc-6032/1762412479704000/
    % \ourmethod{} & 85.8 & 84.3 & 80.6 & 79.2 & 83.7 & 83.3 & 76.4 & 76.0 & 86.4 & 83.2 & 80.1 & 77.0 & 87.4 & 86.4 & 85.4 & 84.4 \\
    % https://waymo.com/open/challenges/detection-3d/results/6f5c8ccc-6032/1762588980398000/
    \rowcolor{blue!10}
    \ourmethod{} & $\underset{\textbf{(+2.8)}}{\textbf{86.7}}$ & $\underset{\textbf{(+2.7)}}{\textbf{85.3}}$ & $\underset{\textbf{(+3.2)}}{\textbf{81.6}}$ & $\underset{\textbf{(+3.1)}}{\textbf{80.2}}$ & $\underset{\textbf{(+1.5)}}{\textbf{84.3}}$ & $\underset{\textbf{(+1.6)}}{\textbf{83.9}}$ & $\underset{\textbf{(+1.7)}}{\textbf{77.1}}$ & $\underset{\textbf{(+1.8)}}{\textbf{76.7}}$ & $\underset{\textbf{(+0.7)}}{\textbf{87.5}}$ & $\underset{\textbf{(+0.5)}}{\textbf{84.7}}$ & $\underset{\textbf{(+1.3)}}{\textbf{81.4}}$ & $\underset{\textbf{(+1.1)}}{\textbf{78.6}}$ & $\underset{\textbf{(+6.0)}}{\textbf{88.2}}$ & $\underset{\textbf{(+5.9)}}{\textbf{87.2}}$ & $\underset{\textbf{(+6.4)}}{\textbf{86.3}}$ & $\underset{\textbf{(+6.3)}}{\textbf{85.4}}$ \\
    \bottomrule
  \end{tabular}%
  }
  \caption{Detailed performance comparison on the WOD \textit{validation} set. We report Average Precision (AP) and Average Precision with Heading (APH) across three agent types at two difficulty levels (L1 and L2). Our method consistently achieves better results across all categories. Baselines include comparable methods that use up to 4 temporal frames. Best results are in \textbf{bold}, annotated with improvements over the runner-up. The ranking metric is \underline{underlined}.}
  \label{tab:wod-comparison-validation}
  % \vspace{-7mm}
\end{table*}

\begin{table}[t]
  \centering
  % \small
  % \resizebox{\textwidth}{!}{%
  \begin{tabular}{@{}l c cc cc@{}}
    \toprule
    \multirow{2}{*}{Method} & \multicolumn{1}{c}{History} & \multicolumn{2}{c}{Overall L1} & \multicolumn{2}{c}{Overall L2} \\
    & Frames & AP & APH & AP & \underline{APH} \\
    \midrule
    $\underset{\text{\cite{zhang2023hednet}}}{\text{HEDNet}}$ & 0 & 82.2 & 80.2 & 76.9 & 75.0 \\
    $\underset{\text{\cite{zhang2023hednet}}}{\text{HEDNet}}$ & 0 & 82.2 & 80.2 & 76.9 & 75.0 \\
    $\underset{\text{\cite{zhou2022centerformer}}}{\text{CenterFormer}}$ & 15 & 82.3 & 80.9 & 77.6 & 76.3 \\
    $\underset{\text{\cite{bevfusion23icra}}}{\text{BEVFusion}}$ & 2 &  82.7 & 81.4 & 77.7 & 76.3 \\
    $\underset{\text{\cite{he2023msf}}}{\text{MSF}}$ & 3 & 83.5 & 82.1 & 78.4 & 77.1\\
    $\underset{\text{\cite{fan2023super}}}{\text{FSD++}}$ & 6 & 83.5 & 82.1 & 78.4 & 77.1\\
    $\underset{\text{\cite{li2023logonet}}}{\text{LoGoNet}}$ & 4 & 83.1 & 81.8 & 78.4 & 77.1\\
    $\underset{\text{\cite{liu2024seed}}}{\text{SEED-L}}$ & 2 & 83.5 & 82.1 & 78.7 & 77.3 \\
    $\underset{\text{\cite{liu2024lion}}}{\text{LION}}$ & 2 &  83.7 & 82.4 & 78.7 & 77.4  \\
    $\underset{\text{\cite{huang2025vadet}}}{\text{VADet}}$ & 15 & 84.1 & 82.8 & 79.4 & 78.2 \\
    $\underset{\text{\cite{agro2025mad}}}{\text{MAD}}$ & 99 & 86.0 & 84.3 & 81.8 & 80.2  \\
    \midrule
    \rowcolor{blue!10}
    \ourmethod{} & 3 & $\underset{\textbf{(+1.5)}}{\textbf{87.5}}$ & $\underset{\textbf{(+1.9)}}{\textbf{86.2}}$ & $\underset{\textbf{(+1.3)}}{\textbf{83.1}}$ & $\underset{\textbf{(+1.6)}}{\textbf{81.8}}$ \\
    \bottomrule
  \end{tabular}%
  \caption{Overall performance comparison on the WOD \textit{test} set. We report the average of Average Precision (AP) and Average Precision with Heading (APH), at two difficulty levels (L1 and L2). Our method achieves state-of-the-art performance using a total of 4 frames (including 3 history frames). Best results are in \textbf{bold}, annotated with improvements over the runner-up. The ranking metric is \underline{underlined}.}
  \label{tab:wod-comparison-test}
  \vspace{-3em}
\end{table}

\begin{figure}[t]
    \centering
    \includegraphics[width=\linewidth]{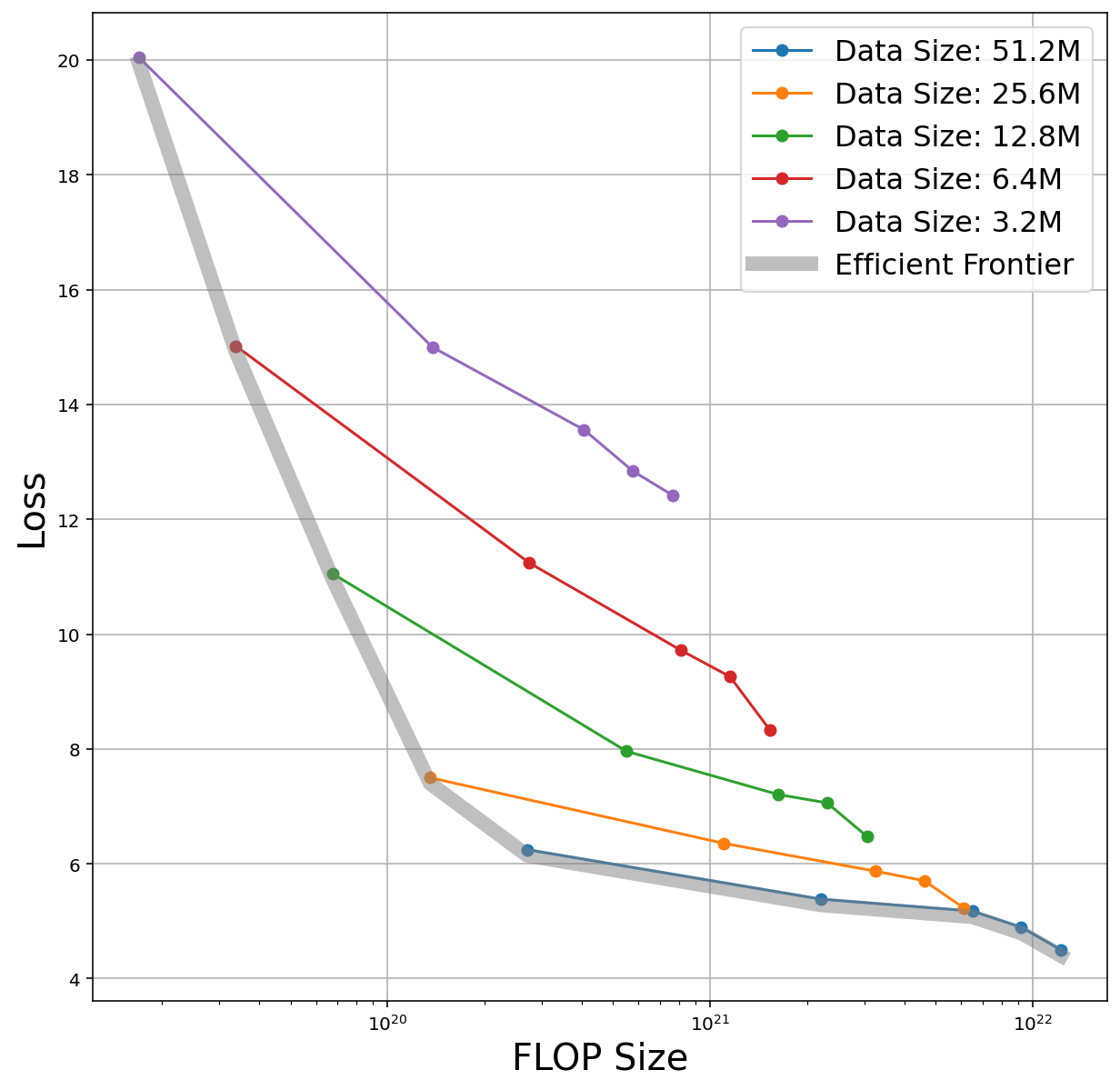}
    \caption{Compute scaling curves. Each dot represents a model size, and each line represents various model sizes training with a given data size. Both large models and larger datasets lead to lower loss. The efficient frontier curve indicates that for a fixed compute FLOPs budget, it is more effective to train a smaller model on a larger dataset than to train a larger model on a small dataset.}
    \label{fig:compute_scaling}
    % \vspace{-2em}
\end{figure}

\subsection{Compute Scaling}
Fig.~\ref{fig:compute_scaling} illustrates the relationship between the total compute, measured in FLOPs (Floating-Point Operations), and the loss. Similar to Fig.~\ref{fig:model_scaling}, each colored line represents a model trained on a fixed dataset size. The trend indicates that as the FLOP budget increases, the model performance improves (i.e., loss decreases), which is consistent with our previous model and data scaling results.

We further visualize the ``efficient frontier," which represents the optimal model performance achievable for a given FLOPs budget. 
This frontier is dominated by the bottom-most curve (representing the largest dataset). 
This empirically demonstrates that the most effective way to utilize a given compute budget is to employ the largest possible dataset.
On the other hand, using a large model on a small dataset (e.g., the rightmost point on the purple line) is highly compute-inefficient, yielding a significantly higher loss compared to what the same compute budget could achieve by using more data.

\section{Evaluation}

We evaluate the largest~\ourmethod{} (\ourmethod{}-483M) on the Waymo Open Dataset (WOD)~\cite{sun2020scalability}, following the mid-training and finetuning methodology described in Sec.~\ref{sec:recipe}. This benchmark consists of over a thousand 20-second scenes containing high-quality LiDAR and camera data from diverse geographies, including approximately 158K training frames and 40K validation frames. The dataset focuses on three major object categories: vehicle, pedestrian, and cyclist. To bridge the domain gap between WOD and our internal data, we finetune our models on the WOD training set before evaluating them on the validation and test sets.

\subsection{Quantitative Results}

Tab.~\ref{tab:wod-comparison-validation} summarizes our results on the WOD validation set. These results demonstrate that \ourmethod{} outperforms prior methods using only causal information (i.e., no future frames) up to 4 frames across all metrics. Notably, it boosts cyclist detection performance by a large margin.

Tab.~\ref{tab:wod-comparison-test} presents a comparison of~\ourmethod{} on the WOD test set against published methods on the leaderboard. We exclude methods that utilize ensembles, future frames, or test-time augmentation. Our results demonstrate that~\ourmethod{} sets a new state-of-the-art in this category. We attribute this success to the benefits of large-scale pre-training, which allows our model to achieve superior performance using significantly fewer temporal frames than complex memory-augmented approaches such as~\cite{agro2025mad}. Furthermore, our method is orthogonal to these memory-augmented strategies, and integrating~\ourmethod{} with such techniques holds strong promise for achieving even greater performance gains in future work.

\subsection{Ablation}
We validate the effectiveness of \ourmethod{} through detailed ablation studies on training recipe, model scaling, mid-training tasks, multi-modal inputs, and temporal context. Unless otherwise specified, these ablations utilize the \ourmethod{}-96M model and are evaluated on a subset of the WOD validation set for computational efficiency.

\subsubsection{Training Recipe}
We first conduct ablations for our training recipe using the \ourmethod{}-96M model.
As shown in Tab.~\ref{tab:recipe_ablation}, the first row establishes a baseline, showing performance when training directly on WOD without pre-training or mid-training.
Subsequent rows add each stage incrementally.
Results indicate that detection performance improves significantly with the addition of pre-training on large-scale auto-labeled data and mid-training on high-quality human-labeled data.

\begin{table}[t!]
\small
\centering
\setlength{\tabcolsep}{4pt} % Tighten spacing
\begin{tabular}{ccc|c} 
\toprule
\multicolumn{3}{c|}{Training Stage} & \multicolumn{1}{c}{L2 APH} \\ 
Pre-training & Mid-training & Finetuning & Overall \\
\midrule
     &       &   $\checkmark$ (From Scratch)    & 66.8       \\
$\checkmark$     &   &  $\checkmark$   & 71.5     \\
$\checkmark$     & $\checkmark$ & $\checkmark$      & 75.7     \\
\bottomrule
\end{tabular}
\caption{Ablation study on training recipe using~\ourmethod{}-96M model. Adding large-scale pre-training and high-quality mid-training stages significantly improves 3D detection in WOD.}
\label{tab:recipe_ablation}
% \vspace{-2em}
\end{table}

\subsubsection{Model Scaling}
Table~\ref{tab:model_size_ablation} presents fine-tuning results on WOD following pre-training across varying model sizes.
To isolate the effects of pre-training scaling, we omit the mid-training stage in these experiments.
Results demonstrate that scaling benefits consistently transfer to WOD performance as model size increases. Fig.~\ref{fig:teaser} further illustrates this trend by plotting performance as a function of FLOPs.

\begin{table}[t!]
\small
\centering
\begin{tabular}{c|cccc}
\toprule
\multicolumn{1}{c|}{Model} & \multicolumn{4}{c}{L2 APH} \\
Parameters & Overall & Vehicle & Pedestrian & Cyclist \\
\midrule
25M & 47.8 & 52.8 & 46.2 & 44.5          \\
96M & 71.5 & 76.3 & 66.8 & 71.6          \\
483M & 78.3 & 81.7 & 75.1 & 78.2          \\
\bottomrule
\end{tabular}
\caption{Ablation study on model scaling. Results indicate that performance gains from scaling consistently transfer to the WOD benchmark as model size increases.}
\label{tab:model_size_ablation}
% \vspace{-2em}
\end{table}

\subsubsection{Mid-Training Tasks} 
We investigate the benefit of introducing additional tasks and supervisions in the mid-training  stage, as described in Sec.~\ref{sec:recipe}. This stage, which uses high-quality human-annotated data, is added on top of a pre-training phase that uses a detection-only task on our vast auto-labeled dataset. The results are presented in Tab.~\ref{tab:task_ablation}. We show that adding extra supervision tasks during mid-training, such as roadgraph prediction and occupancy prediction, progressively improves L2 APH on the WOD validation set.

% https://screenshot.googleplex.com/3hWbpmHzfREzZqd
\begin{table}[t!]
\small
\centering
\setlength{\tabcolsep}{4pt} % Tighten spacing
\begin{tabular}{ccc|c} 
\toprule
\multicolumn{3}{c|}{Midtraining Task} & \multicolumn{1}{c}{L2 APH} \\ 
Detection & Roadgraph Pred. & Occupancy Pred. & Overall \\
\midrule
$\checkmark$     &       &       & 75.7       \\
$\checkmark$     & $\checkmark$   &   & 76.3     \\
$\checkmark$     & $\checkmark$ & $\checkmark$      & 76.6     \\
\bottomrule
\end{tabular}
\caption{Ablation study on mid-training tasks using~\ourmethod{}-96M model. Adding extra tasks during mid-training leads to improved 3D detection performance in WOD.}
\label{tab:task_ablation}
% \vspace{-2em}
\end{table}

\subsubsection{Multi-Modal Inputs} 
Tab.~\ref{tab:input_ablation} presents 3D detection performance on the WOD validation set as we incrementally add input modalities (LiDAR, camera, and surfel map prior) during fine-tuning on WOD. The results indicate that each additional modality improves overall performance, as measured by L2 APH. 

\begin{table}[t!]
\small
\centering
\setlength{\tabcolsep}{4pt} % Tighten spacing
\begin{tabular}{ccc|c} 
\toprule
\multicolumn{3}{c|}{Input Modality} & \multicolumn{1}{c}{L2 APH} \\ 
LiDAR & Camera & Surfel & Overall \\
\midrule
$\checkmark$     &       &       & 74.9      \\
$\checkmark$     & $\checkmark$      &        & 75.7     \\
$\checkmark$     & $\checkmark$      &  $\checkmark$       & 76.1     \\
\bottomrule
\end{tabular}
\caption{Ablation study on input modalities using~\ourmethod{}-96M model. Adding camera and surfel inputs leads to improved 3D detection performance in WOD.}
\label{tab:input_ablation}
% \vspace{-2em}
\end{table}

\begin{figure*}[ht!]
    \centering
    \begin{subfigure}{0.495\textwidth}
        \includegraphics[width=\textwidth, height=5.78cm]{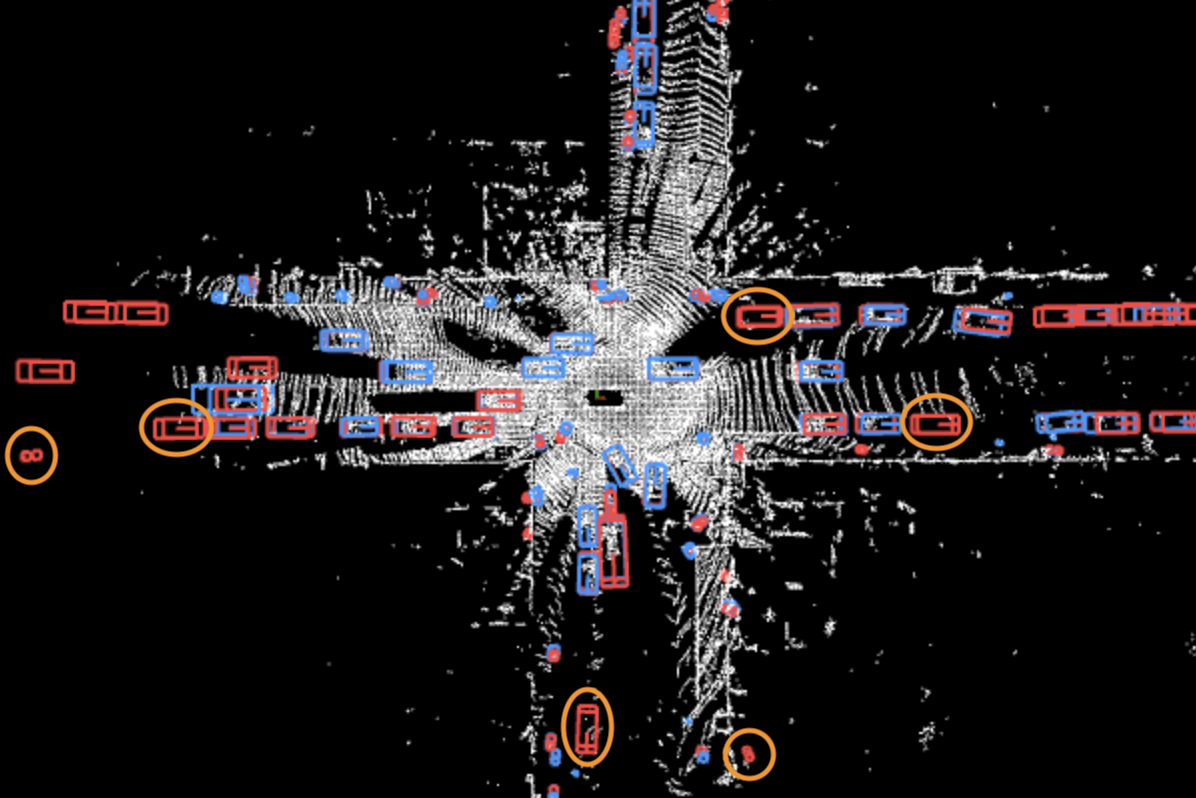}
    \end{subfigure}
    \hfill
    \begin{subfigure}{0.495\textwidth}
        \includegraphics[width=\textwidth, height=5.78cm]{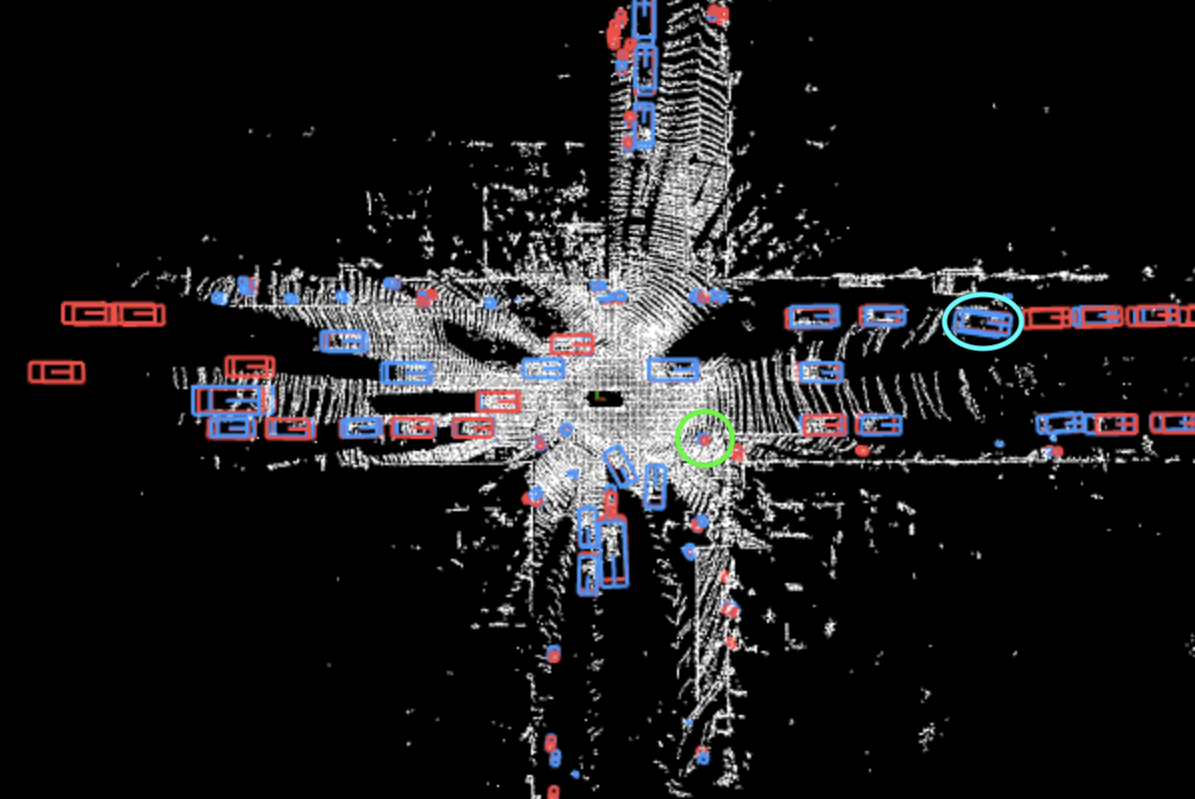}
    \end{subfigure}
    \caption{Qualitative comparison of the \ourmethod{}-96M (left) and \ourmethod{}-483M (right) models, pre-trained on the full dataset and finetuned on the WOD validation set. Ground truth boxes are shown in \textcolor{blue}{blue}, and predictions (confidence $>$ 0.2) are in \textcolor{red}{red}. Compared to the smaller model, the larger model (right) demonstrates superior performance: it successfully detects a pedestrian at the crosswalk (\textcolor{green}{green}), yields more accurate bounding box prediction (\textcolor{cyan}{cyan}), and produces fewer false positives (\textcolor{orange}{orange}).}
    \label{fig:qualitative_example_model}
    % \vspace{-1em}
\end{figure*}

\begin{figure*}[ht!]
    \centering
    \begin{subfigure}{0.495\textwidth}
        \includegraphics[width=\textwidth, height=6.10cm]{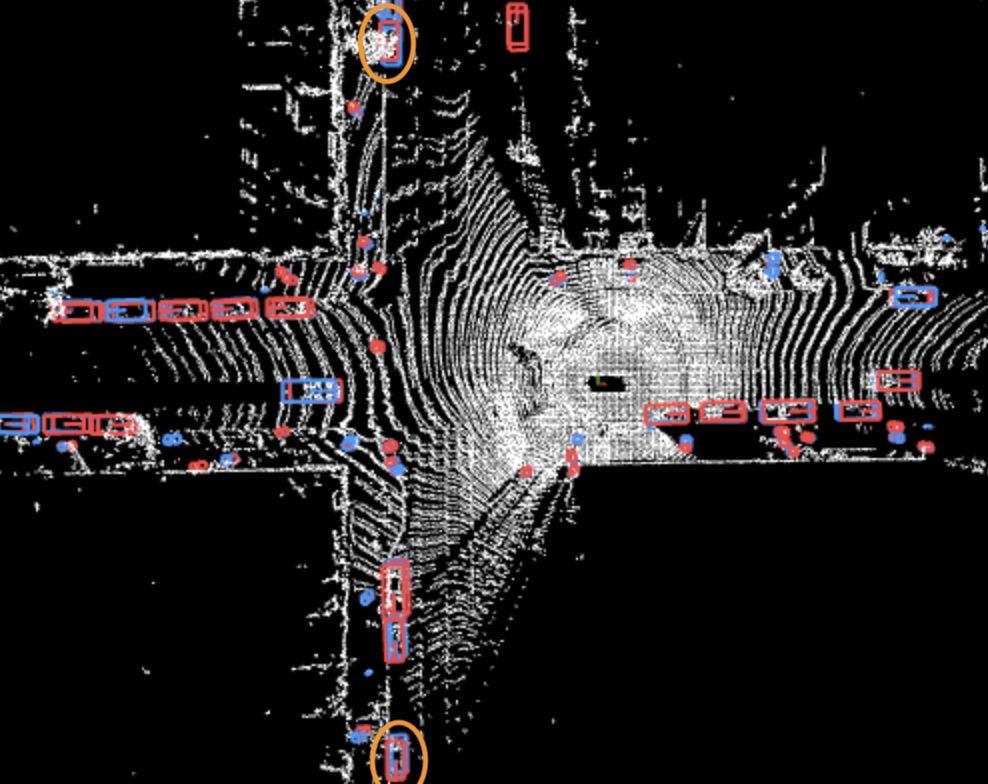}
    \end{subfigure}
    \hfill
    \begin{subfigure}{0.495\textwidth}
        \includegraphics[width=\textwidth, height=6.10cm]{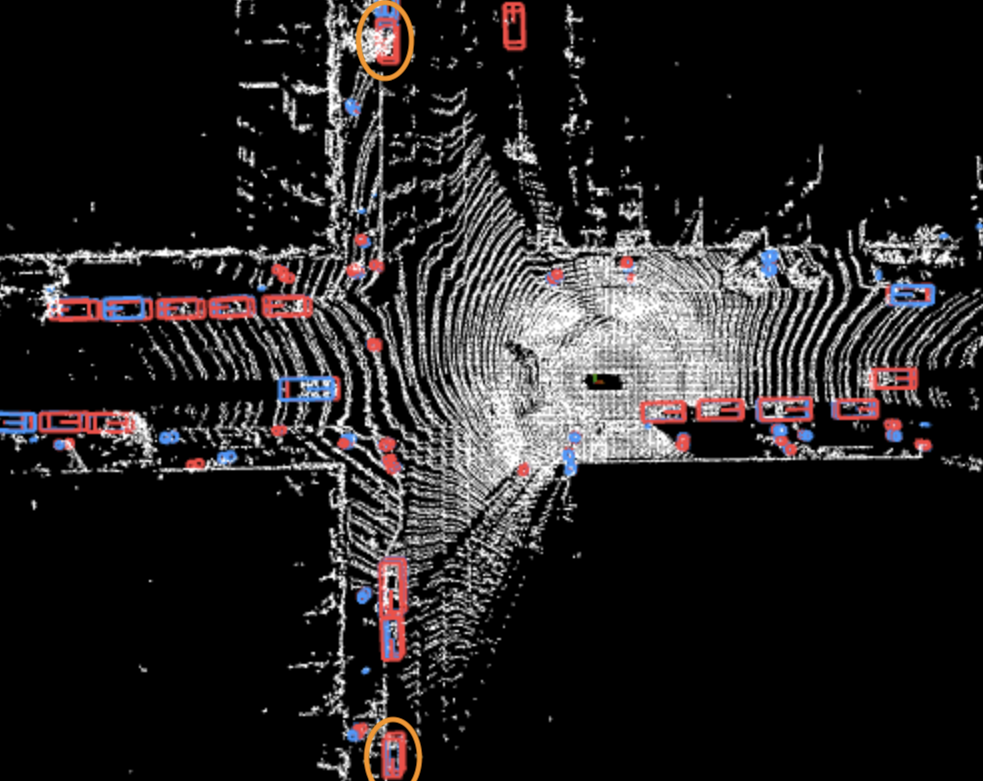}
    \end{subfigure}
    \caption{Qualitative comparison of the \ourmethod{}-483M model pre-trained on the 12.8M dataset (left) vs. the full dataset (right) and finetuned on WOD. Ground truth boxes are shown in \textcolor{blue}{blue}, and predictions (confidence $>$ 0.2) are in \textcolor{red}{red}. The model trained on more examples exhibits better quality detections, especially in long range (highlighted in \textcolor{orange}{orange}).}
    \label{fig:qualitative_example_data}
    % \vspace{-1em}
\end{figure*}

\subsubsection{Temporal Context} 
\label{sec:temporal_context}
Tab.~\ref{tab:temporal_ablation} details the impact of increasing the temporal context when fine-tuning a LiDAR-only model on the WOD validation set.
The results demonstrate a consistent improvement in performance across all object categories as the context window expands up to four frames, which aligns with the temporal context used during the pre-training and mid-training stages.
Beyond this point, the overall improvements begin to plateau.
Notably, we observe a divergence: performance for vehicle and pedestrian types improves slightly, while cyclist detection performance degrades.

To balance between computational efficiency and performance, we select a temporal context of four frames for~\ourmethod{}.

\begin{table}[t!]
\small
\centering
\begin{tabular}{c|cccc}
\toprule
\multicolumn{1}{c|}{Number of} & \multicolumn{4}{c}{L2 APH} \\
Input Frames & Overall & Vehicle & Pedestrian & Cyclist \\
\midrule
2 & 73.3 & 75.9 & 69.2 & 74.7          \\
4 & 75.7 & 78.8 & 71.7 & 76.5          \\
6 & 76.1 & 80.0 & 72.1 & 76.3          \\
8 & 76.4 & 80.7 & 72.3 & 76.1          \\
\bottomrule
\end{tabular}
\caption{Ablation study on the number of input frames using the \ourmethod{}-96M model. Finetuning performance consistently improves up to 4 input frames, aligning with the temporal context used during the pre-training and mid-training stages. Beyond 4 frames, we observe a divergence in performance across different object categories.}
\label{tab:temporal_ablation}
% \vspace{-2em}
\end{table}

\subsubsection{Model Distillation}
During distillation, we employ the best~\ourmethod{} model as teacher, and the student model is the smallest 25M variant. Using our mid-training dataset, the student undergoes a two-stage training process: 160,000 steps of distillation using teacher-generated targets, followed by 40,000 steps of fine-tuning on human-annotated labels. This distilled student achieves a 4.7\% improvement in overall L2 APH compared to a baseline trained exclusively on human-annotated labels for the full 200,000 steps.

\subsection{Qualitative Comparison}
We present two pairs of qualitative examples to demonstrate the effectiveness of scaling in terms of model parameters and data sizes. More examples can be found in Appendix~\ref{appendix:qualitative_examples}.

\subsubsection{Model Scaling}
\label{sec:qualitative_model_scaling}
Fig.~\ref{fig:qualitative_example_model} provides a qualitative comparison between a smaller and a larger model pre-trained on full dataset and finetuned on WOD. The larger model demonstrates superior performance: it predicts box locations more accurately (cyan circle), better handles challenging pedestrian objects with few points (green circle), and produces fewer false positives (orange circles).

\subsubsection{Data Scaling}
\label{sec:qualitative_data_scaling}
Fig.~\ref{fig:qualitative_example_data} provides a qualitative comparison of the same~\ourmethod{}-483M model pre-trained on small versus large datasets and finetuned on WOD. The model trained on the larger dataset demonstrates better performance, by predicting more accurate box locations in longer range with sparser points. As highlighted by the orange circles, its predictions (red) align better with the groundtruth (blue).
\section{Conclusion}
We present \ourmethod{}, a comprehensive study systematically analyzing the impact of scale on 3D perception for autonomous driving. Our model accepts multi-modality inputs from diverse sensors, including LiDAR, radar, camera, and surfel map prior. These sensor data are jointly processed in BEV space using a sparse window Transformer backbone. By training models with up to 500 million parameters on 50 million driving examples, a total compute budget exceeding 10 ZettaFLOPs, we establish empirical scaling trends that relate performance to data volume, model size, and compute. The resulting model sets a new state-of-the-art on the Waymo Open Dataset by a substantial margin. Our findings provide strong evidence that large-scale training, particularly with multi-modal data and multi-task training, is a pivotal direction for advancing perception model capabilities for autonomous driving. 
% \input{sec/9_impact_statement}
% \newpage
\bibliographystyle{icml2026}
\bibliography{main}

\clearpage
\setcounter{page}{1}
% \maketitlesupplementary

\appendix

\begin{table*}[t!]
\scriptsize
\centering
\begin{tabular}{c|c|cccc|ccc|c}
\toprule
\multirow{2}{*}{Stage} & \multirow{2}{*}{Dataset}       & \multicolumn{4}{c|}{Inputs}     & \multicolumn{3}{c|}{Tasks}        &  \multirow{2}{*}{Learning Rate} \\
                       &                                & LiDAR & Radar & Camera & Surfel & Detection & Roadgraph & Occupancy &                         \\ \midrule
Pre-training           & Internal full  & $\checkmark$ (4 frames)     & $\checkmark$     & $\checkmark$      &        & $\checkmark$         &           &           & Constant                      \\
Mid-training           & Internal human-annotated & $\checkmark$ (4 frames)     & $\checkmark$     & $\checkmark$      & $\checkmark$      & $\checkmark$         & $\checkmark$         & $\checkmark$         & Cosine decay                       \\
Finetuning             & WOD                            & $\checkmark$ (4 frames)    &       & $\checkmark$      & $\checkmark$      & $\checkmark$         &           &           & Cosine decay                      \\ \bottomrule
\end{tabular}
\caption{The~\ourmethod{} training recipe consists of three sequential stages: pre-training a detection model on the large-scale internal dataset; mid-training on a human-annotated dataset using a multi-task objective and all input features; and finally, finetuning a detection-only model on WOD. The surfel input is derived from LiDAR and camera inputs, as described in Sec.~\ref{sec:data}.}
\label{tab:recipe_summary}
\end{table*}

\section{Additional Model Details}
Our backbone is based on the SWFormer architecture~\cite{sun2022swformer}, which processes input features through a 5-scale sequence of Transformer blocks. These blocks efficiently generate feature maps at different granularities, using relative strides of $[1, 2, 4, 16, 32]$ across the sequence.

We initialize our smallest model using the implementation details reported in SWFormer: a channel size (hidden dimension) of $128$, an MLP ratio (feed-forward ratio) of $2$, a head size of $8$, and layer counts of $[2, 3, 2, 3, 2]$ for the five scales, respectively. From this baseline, we progressively scale the model size up to approximately $500$ million parameters, as detailed in Tab.~\ref{tab:model_scaling}.

The occupancy prediction task classifies voxels into a finite set of semantic labels. The task head is composed of a $3 \times 3$ convolutional layer and an MLP that processes backbone features to output a single class per voxel. The roadgraph prediction task head follows a similar design. We render map elements, such as lanes and road boundaries, in bird's-eye view (BEV) to define per-voxel supervision targets. We adopt simple, lightweight heads for both tasks; this design facilitates robust multi-task mid-training and effectively encourages the backbone to generalize across distinct perception domains.

\section{Additional Training Details}
Tab.~\ref{tab:recipe_summary} provides a detailed comparison of our three-stage training recipe. We select inputs and tasks for each stage based on the scale and quality of the training dataset.

\section{Quantitative Comparison on Rare Classes}
Tab.~\ref{tab:model_size_ablation} demonstrates the significant improvement in L2 APH on the WOD dataset achieved by scaling model parameters. Given that the object classes in WOD are primarily focused on common classes, we extended our evaluation to a set of challenging object classes (e.g., stroller, animal, and traffic cone) using an internal dataset. The results confirm the effectiveness of scaling on these difficult categories: the mAP improved by 5.5\% from~\ourmethod{}-25M to~\ourmethod{}-96M, and by 7.9\% from~\ourmethod{}-96M to~\ourmethod{}-483M. 
Furthermore, in a more rigorous benchmark focused on rare and hard cases,~\ourmethod{}-483M achieved a 25\% error rate reduction over~\ourmethod{}-96M across various AP and recall metrics.
These findings collectively suggest that the benefits of scaling extend beyond common object types, generalizing to more infrequent and challenging rare classes.

\section{Additional Ablation Studies}
In this section, we provide additional ablation studies on temporal context during mid-training and finetuning, as well as perception range during training and inference.

\begin{figure}[t!]
    \centering
    \includegraphics[width=0.95\linewidth]{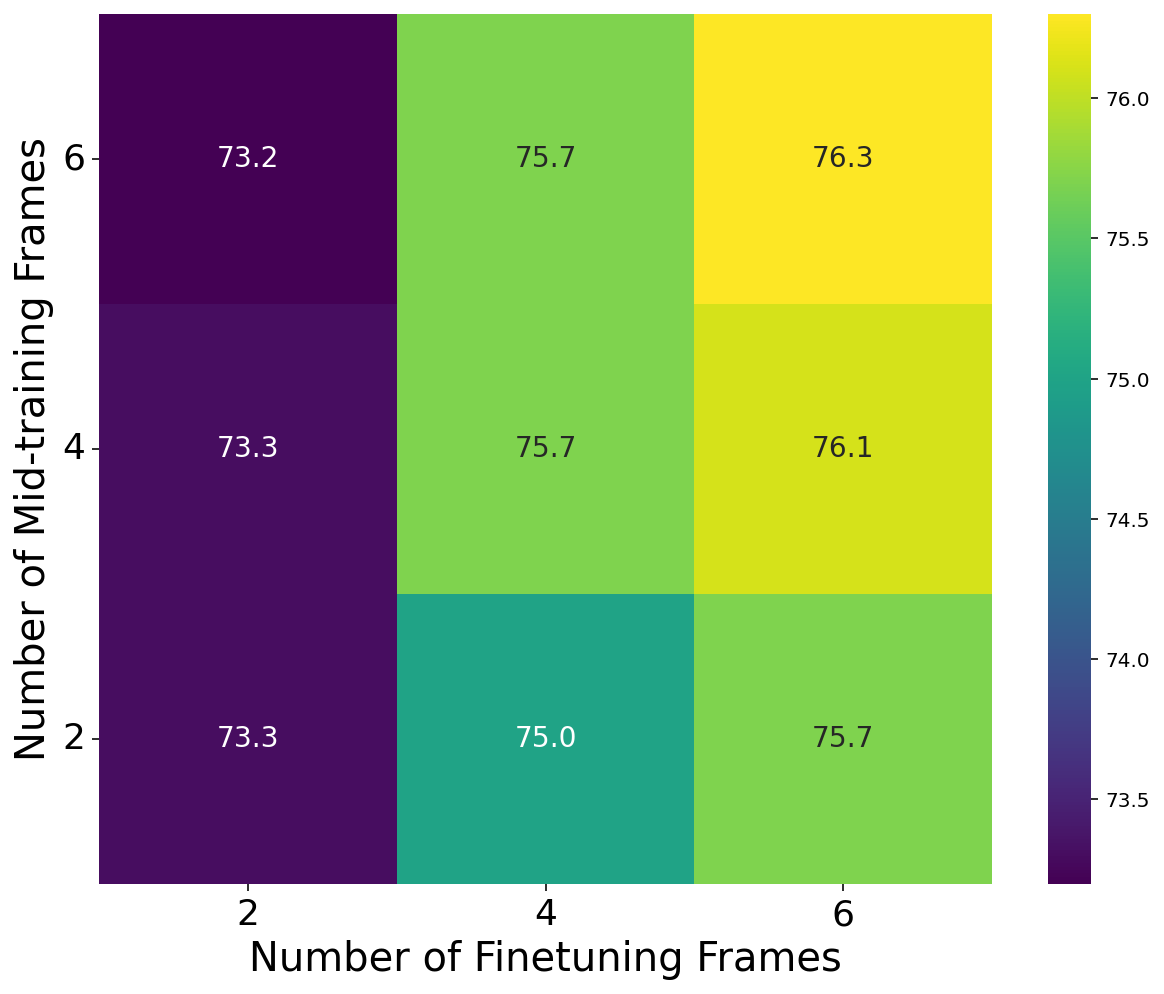}
    \caption{Temporal context ablation across mid-training and finetuning. The results reveal that Overall L2 APH consistently improves as the number of finetuning frames increases, regardless of the mid-training frames at $(2, 4, 6)$. Furthermore, longer context in mid-training offers limited benefit when finetuning uses fewer frames.}
\label{fig:temporal_ablation_mid_training}
\end{figure}

\subsection{Temporal Context during Mid-training}
To assess the impact of temporal context length during mid-training, we extend the ablation study presented in Sec.~\ref{sec:temporal_context}, which fixed the mid-training temporal frames to 4. Fig.~\ref{fig:temporal_ablation_mid_training} summarizes the Overall L2 APH results across different combinations of mid-training and finetuning temporal frames.

We find that the Overall L2 APH consistently improves as the number of finetuning frames increases, irrespective of the mid-training configuration. While the peak performance is achieved using 6 temporal frames, we observe that longer context in mid-training offers limited benefit when finetuning uses fewer frames. For example, mid-training with 6 frames yields no performance gain over 4 frames when the model is subsequently finetuned using only 4 frames. This behavior is further supported by the 2-frame finetuning results, which exhibit highly similar performance regardless of the number of frames used during mid-training.

\begin{table*}[t!]
\centering
\begin{tabular}{c|cc}
\toprule
\multirow{2}{*}{Perception Range (Inference)}  & \multicolumn{2}{c}{Overall L2 APH} \\
 & Range [0, 30) & Range [30, 50) \\
\midrule
75 & 86.5 & 74.7          \\
50 & 86.8 & 74.8     \\
30 & 86.8 & 0.0    \\
\bottomrule
\end{tabular}
\caption{Ablation study on perception range during inference, by training the~\ourmethod{}-96M model at 75-meter range and reducing the range during inference. 
The results show that performance is comparable within the available detection range when the inference distance is reduced, validating the model's robustness to varying ranges.}
\label{tab:range_ablation}
\end{table*}

\begin{table*}[t!]
\centering
\begin{tabular}{c|cc}
\toprule
\multirow{2}{*}{Perception Range (Training)}  & \multicolumn{2}{c}{Overall L2 APH} \\
 & Range [0, 30) & Range [30, 50) \\
\midrule
30 & 86.7 & 0.0          \\
50 & 86.7 & 74.3     \\
75 & 86.5 & 74.7    \\
\bottomrule
\end{tabular}
\caption{Ablation study on perception range, by training the~\ourmethod{}-96M model at different ranges and evaluating the model performance at two ranges. 
The results show that increasing the range during training does not sacrifice performance in smaller ranges.}
\label{tab:range_ablation_training}
\end{table*}

\subsection{Perception Range}
To validate the scalability of our model with respect to perception range during inference, we evaluated the~\ourmethod{}-96M model, originally trained with the WOD range setting at 75 meters, by modifying the inference range to smaller values. The results, reported in Tab.~\ref{tab:range_ablation}, demonstrate that performance remains comparable within the available detection range when the inference distance is reduced. When the detection range is truncated to 30 meters, the Overall L2 APH metric for the $[30, 50)$ meter range is no longer applicable, as expected.

We further investigate the effect of varying the perception range during training and evaluate the resulting detection performance at two different ranges. The results in Tab.~\ref{tab:range_ablation_training} indicate that increasing the perception range during training does not sacrifice performance in smaller ranges.

These two ablations, conducted during both inference and training, collectively validate that our model is robust and scalable across different perception range settings.

\section{Additional Qualitative Examples}
\label{appendix:qualitative_examples}
We present additional qualitative comparisons from model scaling and data scaling as follows.

\subsection{Model Scaling}
\begin{figure*}[t!]
    \centering
    \begin{subfigure}{0.495\textwidth}
        \includegraphics[width=\textwidth]{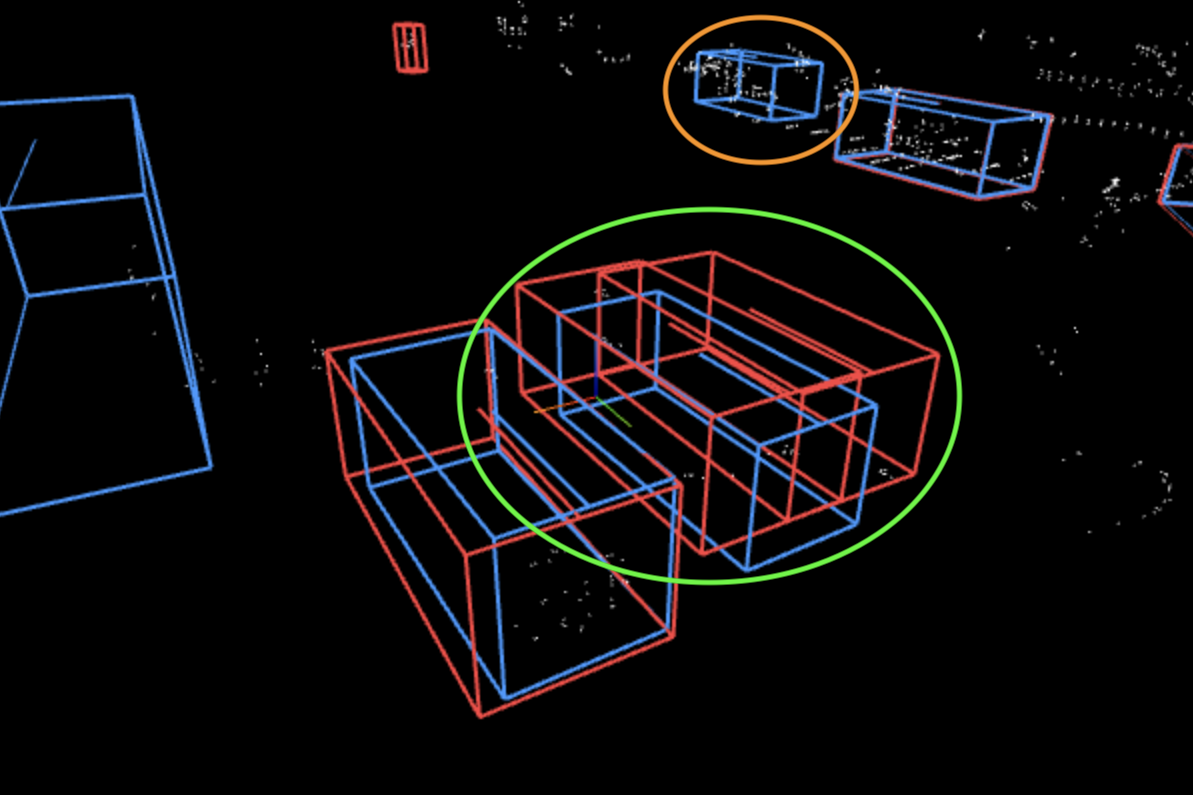}
    \end{subfigure}
    \hfill
    \begin{subfigure}{0.495\textwidth}
        \includegraphics[width=\textwidth]{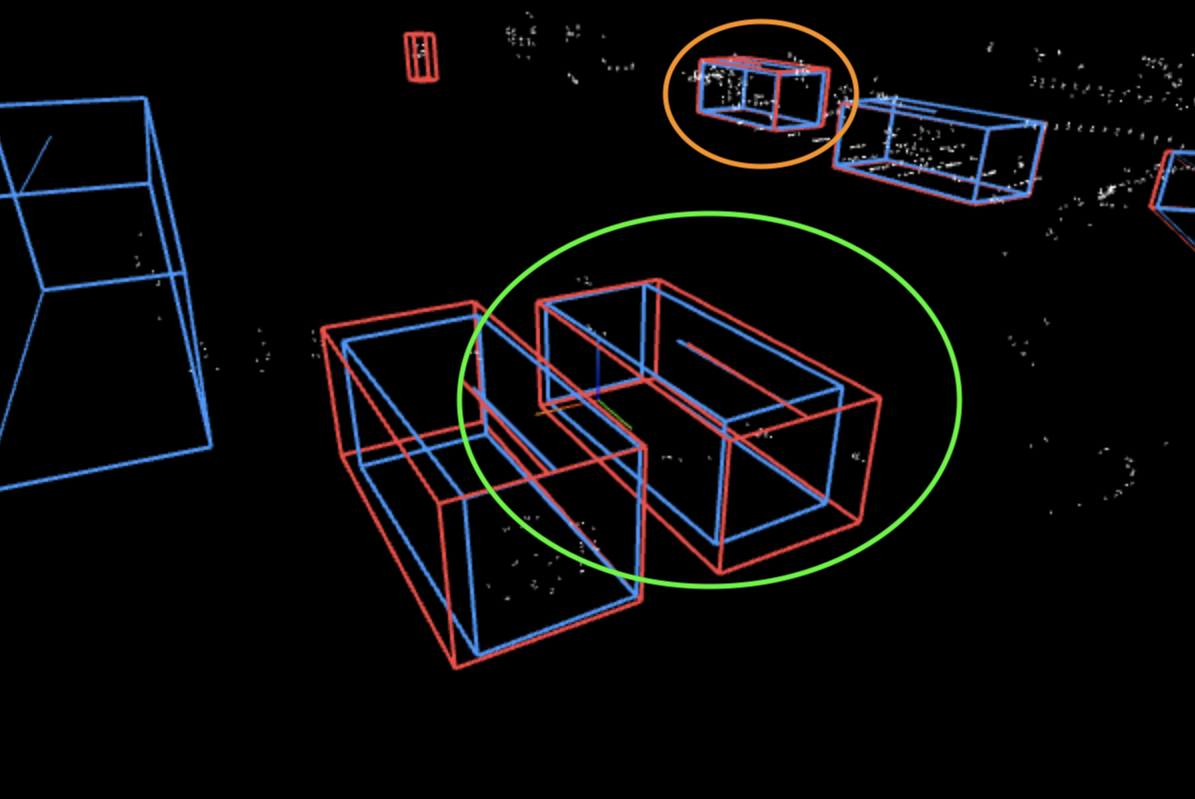}
    \end{subfigure}
    (a) Detection with sparse points.
    \vspace{1.5mm}
    \\
    \begin{subfigure}{0.495\textwidth}
        \includegraphics[width=\textwidth]{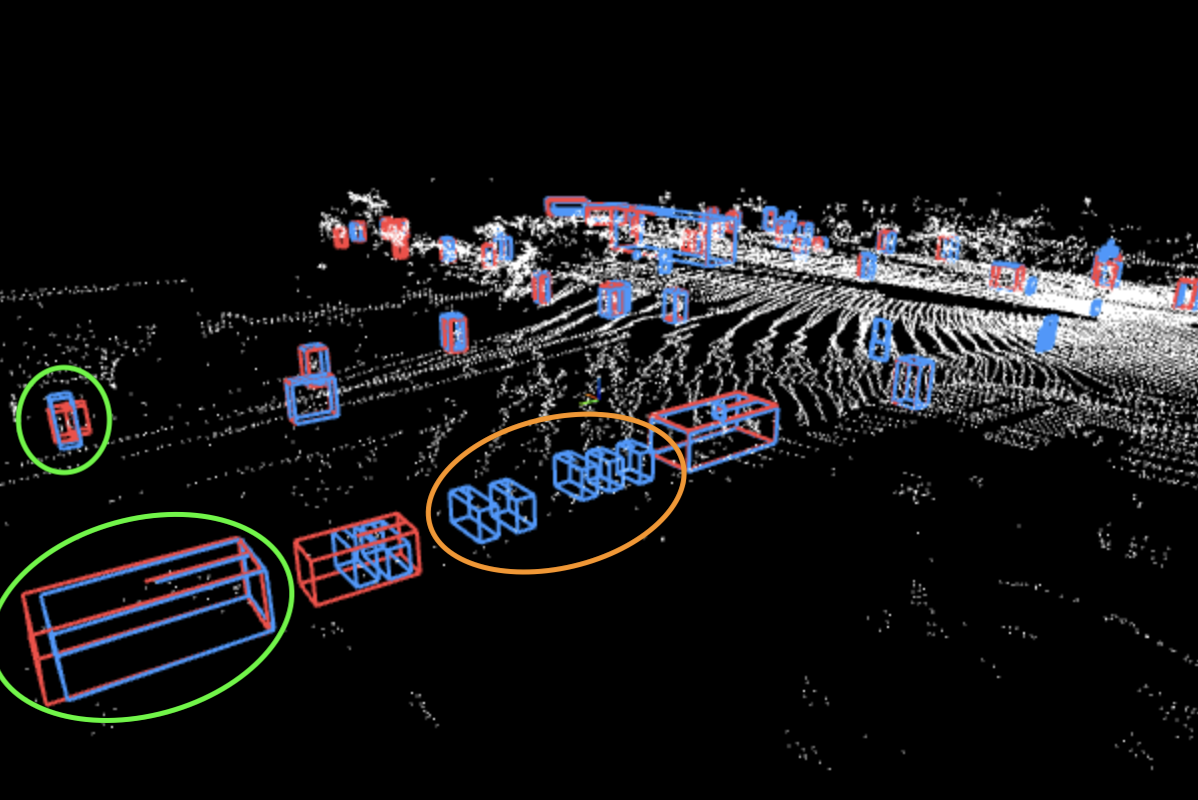}
    \end{subfigure}
    \hfill
    \begin{subfigure}{0.495\textwidth}
        \includegraphics[width=\textwidth]{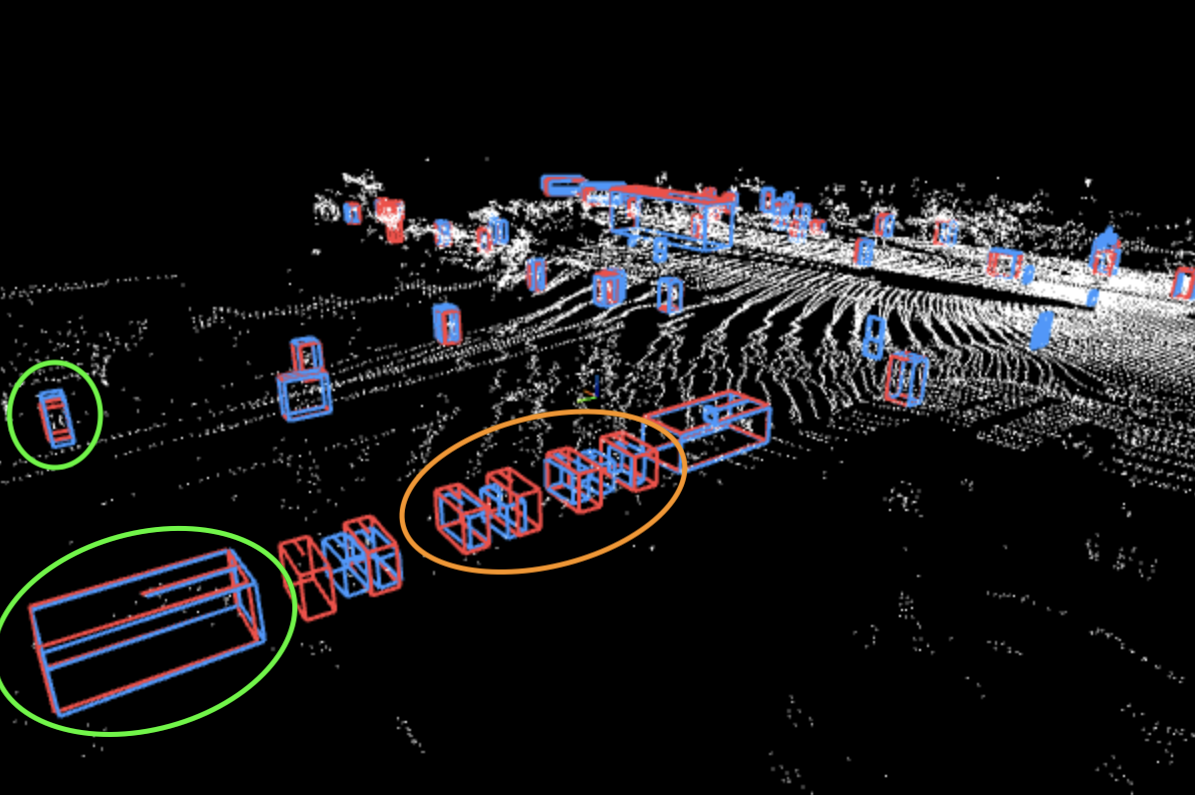}
    \end{subfigure}
    (b) Crowded scene detection.
    \vspace{1.5mm}
    \\
    \begin{subfigure}{0.495\textwidth}
        \includegraphics[width=\textwidth]{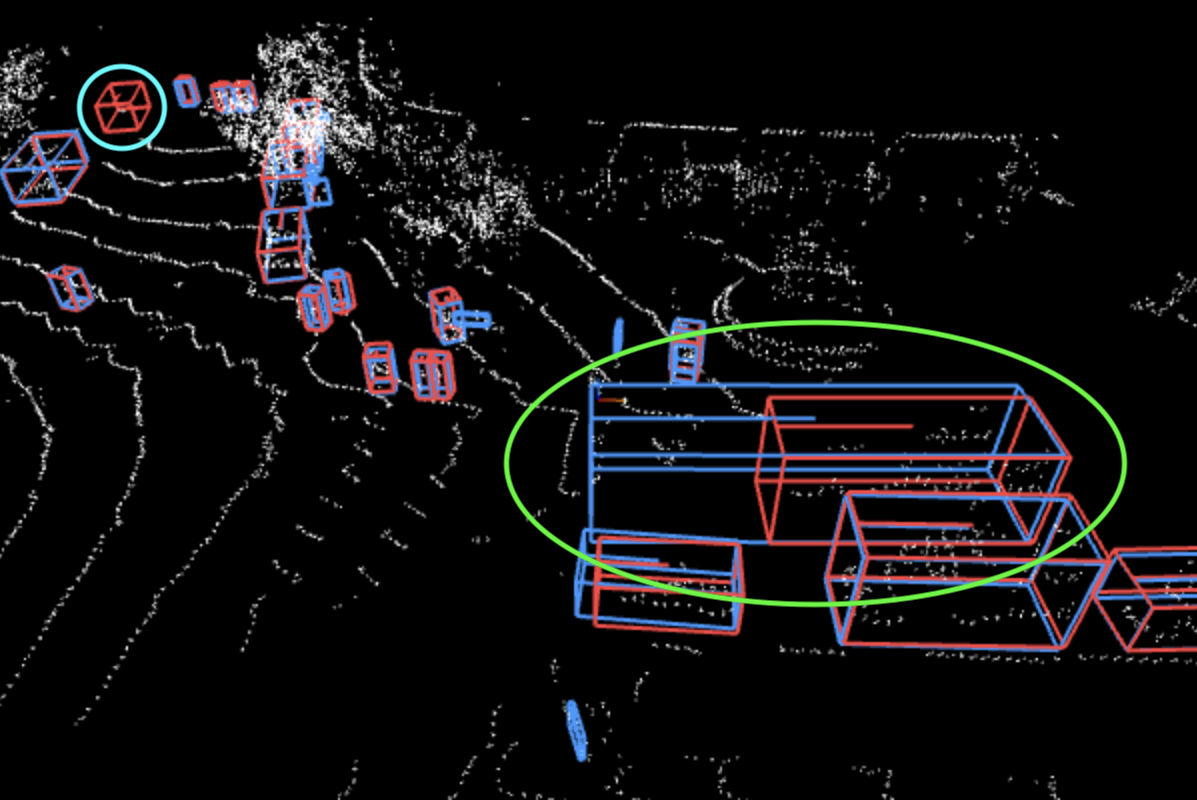}
    \end{subfigure}
    \hfill
    \begin{subfigure}{0.495\textwidth}
        \includegraphics[width=\textwidth]{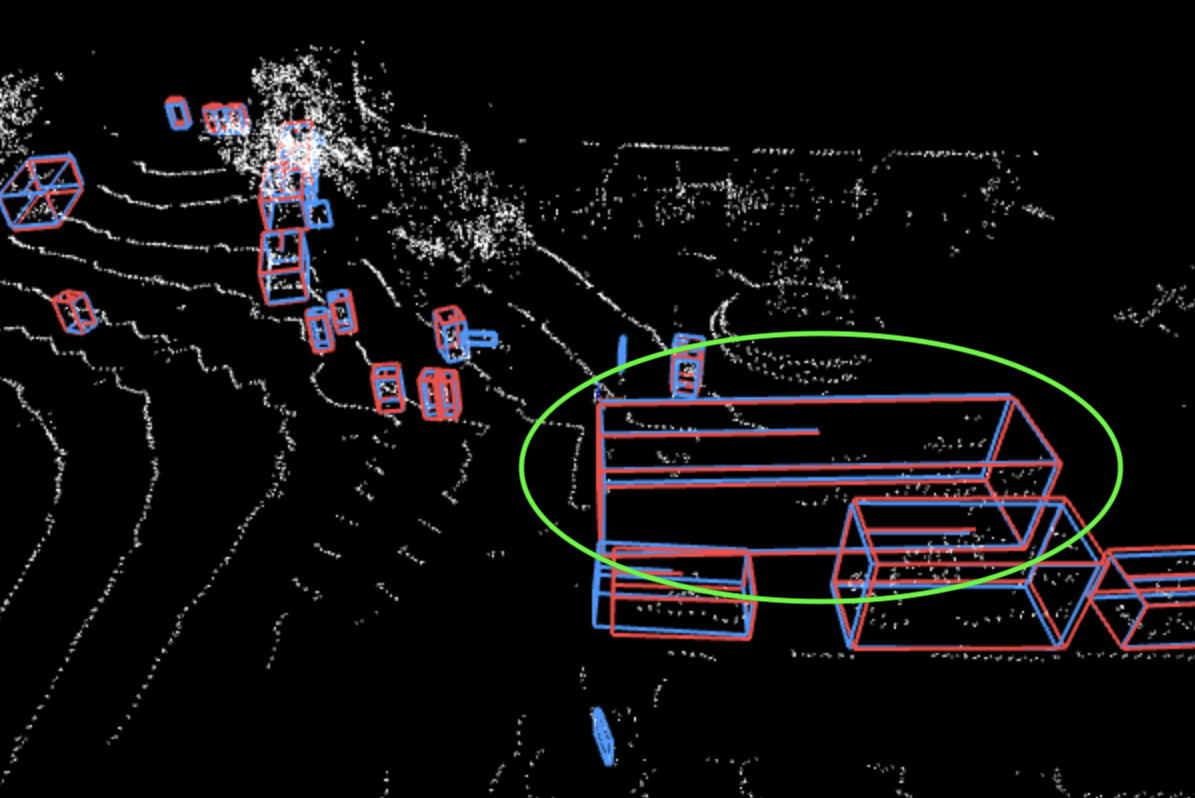}
    \end{subfigure}
    (c) Large object detection.
    \caption{Qualitative comparison of the \ourmethod{}-96M (left column) and \ourmethod{}-483M (right column) models, pre-trained on the full internal dataset and finetuned on WOD. Ground truth boxes are shown in \textcolor{blue}{blue}, and predictions (confidence $>$ 0.2) are in \textcolor{red}{red}.
    In all three examples, the larger model  demonstrates superior performance, achieving higher recall (\textcolor{orange}{orange}) and predicting more accurate location and size (\textcolor{green}{green}), in challenging scenarios involving sparse points, crowded scene, and large objects. In contrast, the smaller model exhibits missed detections (\textcolor{orange}{orange}), inaccurate predictions (\textcolor{green}{green}), and false positives (\textcolor{cyan}{cyan}).
    }
    \label{fig:qualitative_example_model_supp1}
\end{figure*}

Fig.~\ref{fig:qualitative_example_model_supp1} presents additional qualitative comparisons between the \ourmethod{}-96M and \ourmethod{}-483M models. Consistent with the observations in Sec.~\ref{sec:qualitative_model_scaling}, the larger model demonstrates better performance. 
Specifically, it achieves higher recall and generates bounding boxes with more accurate location and size in challenging scenarios, including those with sparse points, crowded scenes, and large objects.

\subsection{Data Scaling}
\begin{figure*}[t!]
    \centering
    \begin{subfigure}{0.495\textwidth}
        \includegraphics[width=\textwidth]{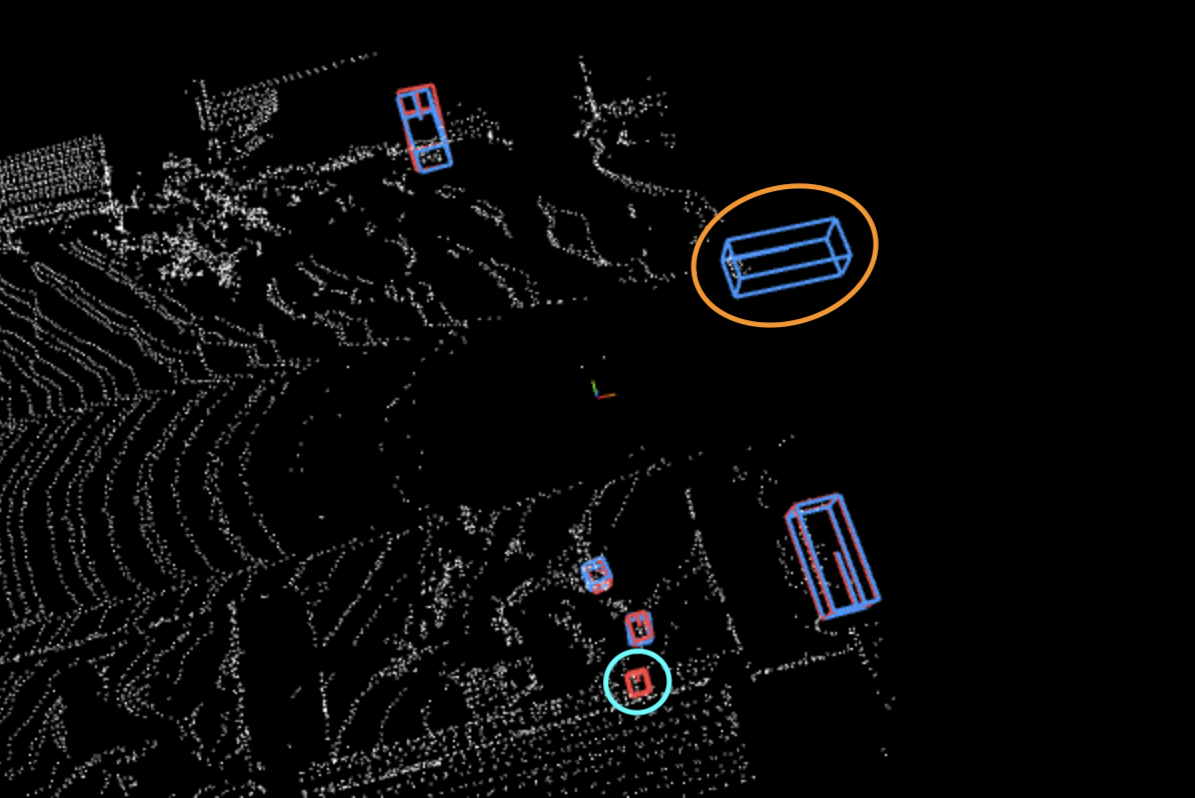}
    \end{subfigure}
    \hfill
    \begin{subfigure}{0.495\textwidth}
        \includegraphics[width=\textwidth]{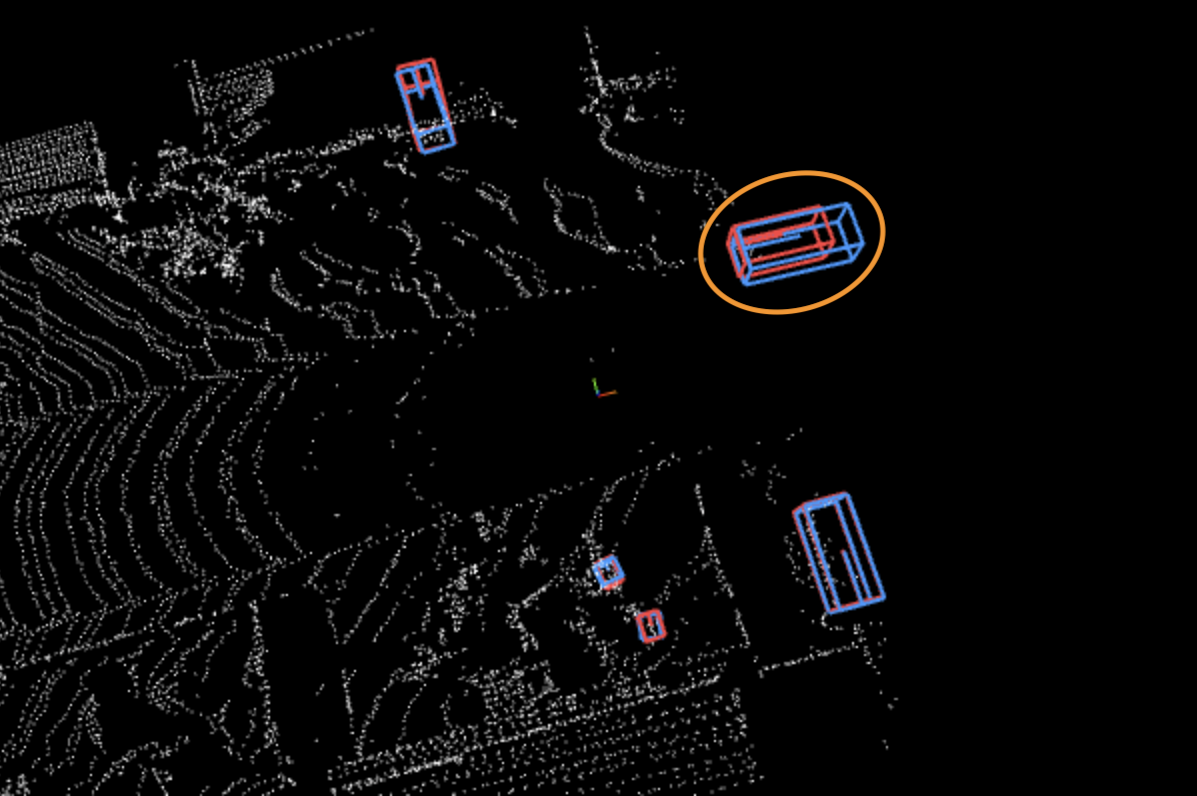}
    \end{subfigure}
    (a) Detection with sparse points.
    \vspace{1.5mm}
    \\
    \begin{subfigure}{0.495\textwidth}
        \includegraphics[width=\textwidth]{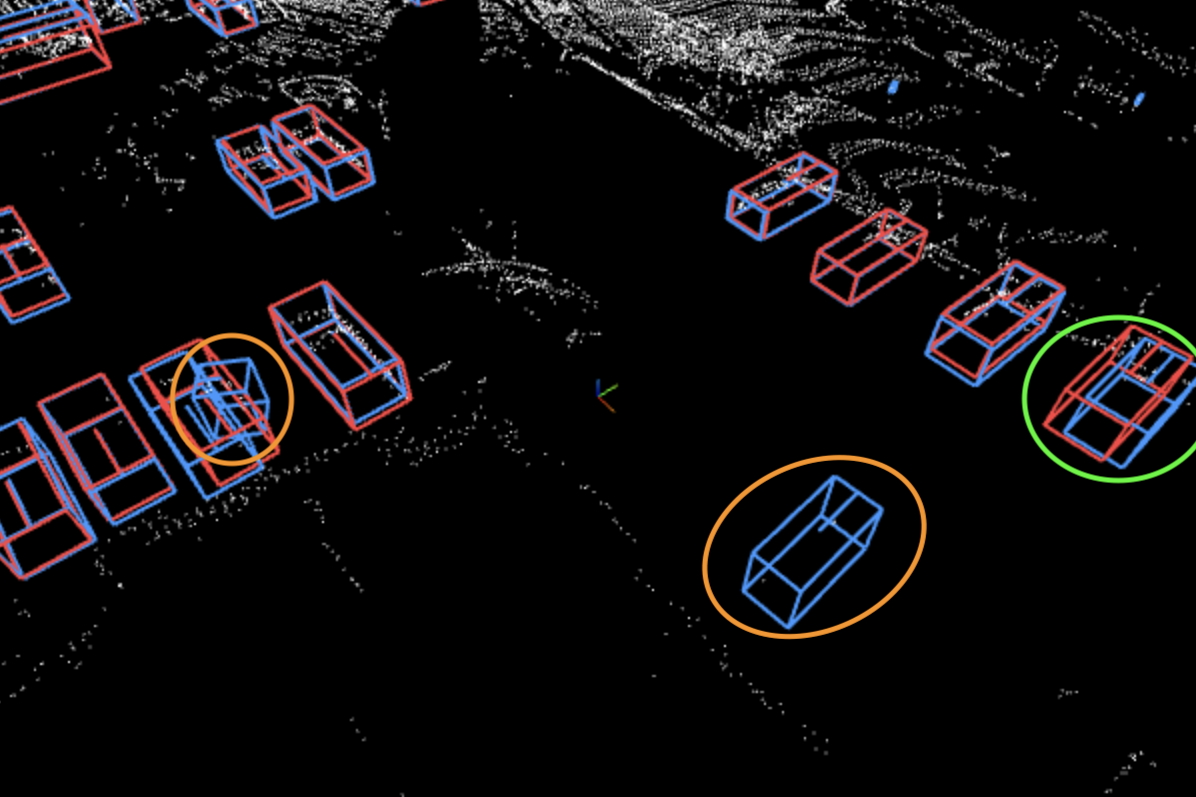}
    \end{subfigure}
    \hfill
    \begin{subfigure}{0.495\textwidth}
        \includegraphics[width=\textwidth]{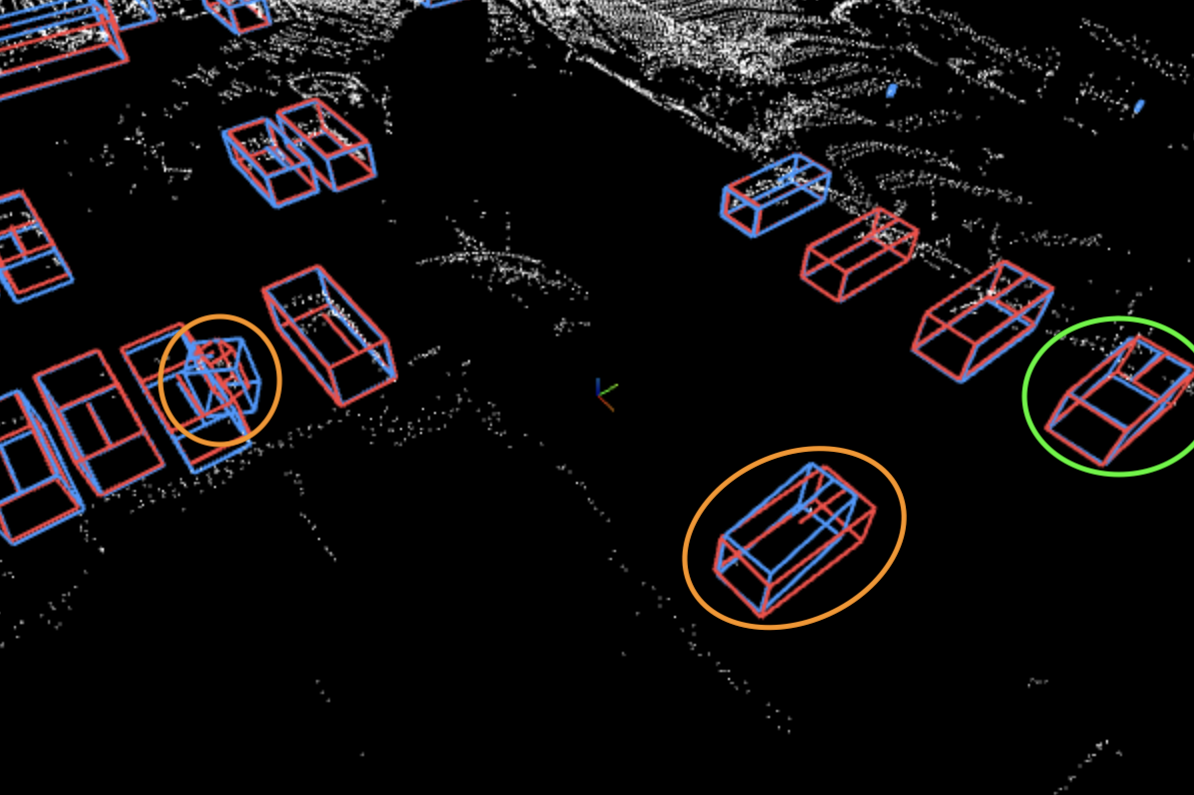}
    \end{subfigure}
    (b) Detection with partial occlusions.
    \vspace{1.5mm}
    \\
    \begin{subfigure}{0.495\textwidth}
        \includegraphics[width=\textwidth]{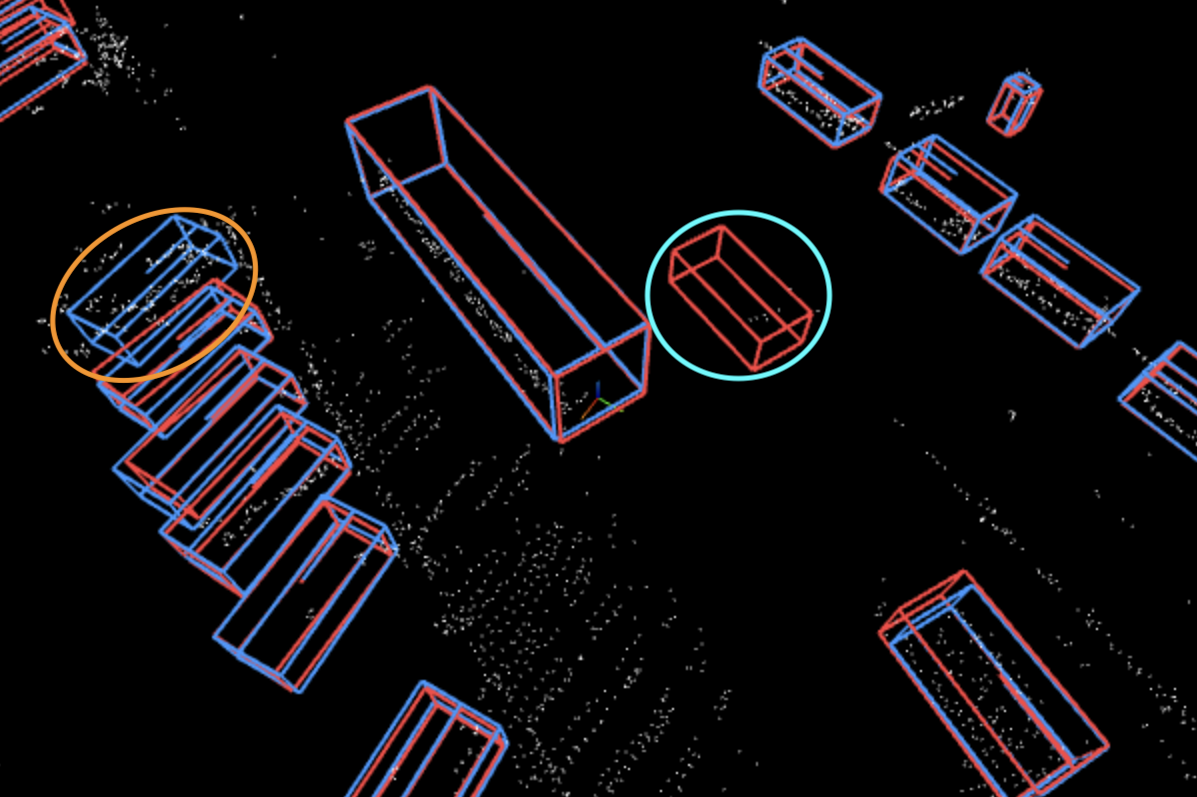}
    \end{subfigure}
    \hfill
    \begin{subfigure}{0.495\textwidth}
        \includegraphics[width=\textwidth]{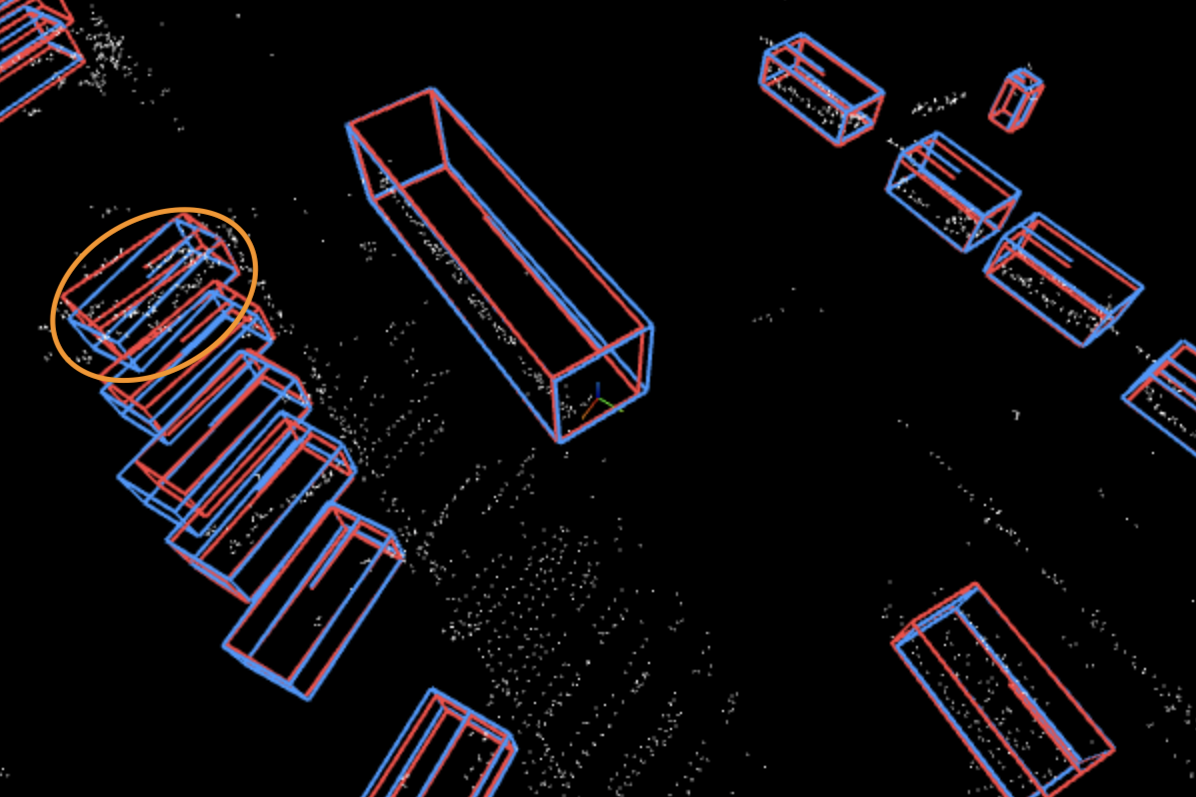}
    \end{subfigure}
    (c) Crowded scene detection.
    \caption{Qualitative comparison of the \ourmethod{}-483M model pre-trained on the 12.8M dataset (left) vs. the full dataset (right) and finetuned on WOD. Ground truth boxes are shown in \textcolor{blue}{blue}, and predictions (confidence $>$ 0.2) are in \textcolor{red}{red}. The model trained on more examples achieves higher recall (\textcolor{orange}{orange}) and more accurate location (\textcolor{green}{green}) in challenging scenarios, involving sparse points, partial occlusions, and crowded scene. In contrast, the smaller model exhibits missed detections (\textcolor{orange}{orange}), inaccurate predictions (\textcolor{green}{green}), and false positives (\textcolor{cyan}{cyan}).
    }
    \label{fig:qualitative_example_data_supp1}
\end{figure*}

Fig.~\ref{fig:qualitative_example_data_supp1} provides additional qualitative comparisons between two instances of the~\ourmethod{}-483M model: one trained on a smaller dataset and the other on the full dataset. Consistent with the observations in Sec.~\ref{sec:qualitative_data_scaling}, the model trained on the larger dataset demonstrates better detection results, particularly in scenarios involving sparse points, partial occlusions, and crowded scenes. In contrast, the counterpart trained on the smaller dataset exhibits a higher rate of false positives and less accurate detections.

\end{document}